\documentclass[wcp]{jmlr}
\jmlrvolume{266}
\jmlryear{2025}
\jmlrworkshop{Conformal and Probabilistic Prediction with Applications}
\jmlrproceedings{PMLR}{Proceedings of Machine Learning Research}

\usepackage{graphicx} 

\usepackage{amssymb}
\usepackage{hyperref}       
\usepackage{url}            
\usepackage{booktabs}       
\usepackage{amsfonts}       
\usepackage{nicefrac}       
\usepackage{microtype}      
\usepackage[x11names]{xcolor}

\usepackage{algorithm}
\usepackage{algcompatible}

\usepackage{dsfont}
\usepackage{amsmath}

\usepackage{cleveref}

\usepackage{appendix}

\usepackage{multirow}
\usepackage{textcomp}

\usepackage{etoolbox}
\makeatletter
\patchcmd{\hyper@makecurrent}{%
    \ifx\Hy@param\Hy@chapterstring
        \let\Hy@param\Hy@chapapp
    \fi
}{%
    \iftoggle{inappendix}{
        \@checkappendixparam{chapter}%
        \@checkappendixparam{section}%
        \@checkappendixparam{subsection}%
        \@checkappendixparam{subsubsection}%
        \@checkappendixparam{paragraph}%
        \@checkappendixparam{subparagraph}%
    }{}%
}{}{\errmessage{failed to patch}}

\newcommand*{\@checkappendixparam}[1]{%
    \def\@checkappendixparamtmp{#1}%
    \ifx\Hy@param\@checkappendixparamtmp
        \let\Hy@param\Hy@appendixstring
    \fi
}
\makeatletter

\newtoggle{inappendix}
\togglefalse{inappendix}

\apptocmd{\appendix}{\toggletrue{inappendix}}{}{\errmessage{failed to patch}}
\apptocmd{\subappendices}{\toggletrue{inappendix}}{}{\errmessage{failed to patch}}

\title{SEMF: Supervised Expectation-Maximization Framework for\\[0.2cm]Predicting Intervals}
\author{
  \Name{Ilia Azizi}\Email{ilia.azizi@unil.ch}\\
  \addr{HEC, University of Lausanne, Lausanne, Switzerland}\\
  \Name{Marc-Olivier Boldi}\Email{marc-olivier.boldi@unil.ch}\\
  \addr{HEC, University of Lausanne, Lausanne, Switzerland}\\
  \Name{Valérie Chavez-Demoulin}\Email{valerie.chavez@unil.ch}\\
  \addr{HEC, University of Lausanne, Expertise Center for Climate Extremes (ECCE), Lausanne, Switzerland}
}

\editor{Khuong An Nguyen, Zhiyuan Luo, Tuwe Löfström, Lars Carlsson and Henrik Boström}

\begin{document}

\maketitle

\begin{abstract}
This work introduces the Supervised Expectation-Maximization Framework (SEMF), a versatile and model-agnostic approach for generating prediction intervals with any ML model. SEMF extends the Expectation-Maximization algorithm, traditionally used in unsupervised learning, to a supervised context, leveraging latent variable modeling for uncertainty estimation. Through extensive empirical evaluation of diverse simulated distributions and 11 real-world tabular datasets, SEMF consistently produces narrower prediction intervals while maintaining the desired coverage probability, outperforming traditional quantile regression methods. Furthermore, without using the quantile (pinball) loss, SEMF allows point predictors, including gradient-boosted trees and neural networks, to be calibrated with conformal quantile regression. The results indicate that SEMF enhances uncertainty quantification under diverse data distributions and is particularly effective for models that otherwise struggle with inherent uncertainty representation.\footnote{Code: \url{https://github.com/Unco3892/semf-paper}}
\end{abstract}


\section{Introduction}
In the evolving field of machine learning (ML), the quest for models able to predict outcomes while quantifying the uncertainty of their predictions is critical. The ability to estimate prediction uncertainty is particularly vital in high-stakes domains such as healthcare \citep{dusenberry2020uncertainehr}, finance \citep{wisniewski2020confirmalfinance}, and autonomous systems \citep{tang2022prediction}, where prediction-based decisions have important consequences. Traditional approaches have primarily focused on point estimates, with little to no insight into prediction reliability. This limitation underscores the need for frameworks that can generate both precise point predictions and robust prediction intervals. Such intervals provide a range within which the true outcome is expected to lie with a fixed probability, offering a finer understanding of prediction uncertainty. This need has spurred research into methodologies that extend beyond point estimation to include uncertainty quantification, thereby enabling more informed decision-making in applications reliant on predictive modeling \citep{ghahramani2015probabilistic}.


In this paper, we introduce the Supervised Expectation-Maximization Framework (SEMF) based on the Expectation-Maximization (EM) algorithm \citep{Dempster1977}. Traditionally recognized as a clustering technique, EM is used for supervised learning in SEMF, allowing for both point estimates and prediction intervals using any ML model (model-agnostic), with a particular focus on uncertainty quantification. While EM has been predominantly used in unsupervised settings, its application to supervised learning represents a principled extension that leverages the algorithm's ability to handle latent variables for uncertainty quantification, a crucial need in modern ML applications. This paper details the methodology behind the framework and proposes a training algorithm based on Monte Carlo (MC) sampling, also used in variational inference for Variational Auto-Encoders (VAEs) \citep{Blei2017, kingma2014vae}. 

Our method, SEMF, differs from prominent supervised EM approaches such as \citet{ghahramani1993em}, which focus on point prediction using Gaussian Mixture Models (GMMs). Unlike VAE-based approaches that optimize the evidence lower bound (ELBO), SEMF directly maximizes the likelihood through importance-weighted sampling, avoiding the need for variational approximations and their potential limitations, such as poor posterior approximation quality when the true posterior $p(z|x,y)$ is multimodal or highly non-Gaussian, and the inherent gap between the ELBO and true likelihood that can compromise uncertainty estimates. Furthermore, SEMF operates in a frequentist paradigm, directly maximizing the likelihood function through iterative EM steps without integrating over posterior distributions. Although SEMF can be extended to a Bayesian framework as its likelihood component, this extension lies beyond the scope of this paper. Finally, SEMF generates representations for latent modalities through specialized models, holding potential for multi-modal data applications.

The remainder of this paper is organized as follows: \autoref{method} details the theory and methodology of SEMF. \autoref{related_work} reviews related work in latent representation learning and uncertainty estimation. \autoref{experimental_setup} describes the experimental setup, including synthetic and benchmark datasets, as well as evaluation metrics. \autoref{results} discusses the results, demonstrating the efficacy of SEMF. Lastly, \autoref{conclusion} concludes the paper, and \autoref{future_work} outlines the limitations and potential research directions.

\section{Method}
\label{method}
This section presents the founding principles of SEMF from its parameters, training, and inference procedure with, at its core, the EM algorithm. Invented to maximize the model likelihood, it builds a sequence of parameters that guarantee an increase in the log-likelihood \citep{wu1983convergence} by iterating between the Expectation (E) and the Maximization (M) steps. In the E-step, one computes

\begin{equation}
\label{eqn:em_e_step}
Q(p | p') = \mathbb{E}_{Z \sim p'(z| x)}\left[ \log p(x, Z) \right] = \int \log p(x,z) p'(z|x) dz,
\end{equation}
where \( p' \) stands for the current estimates, \( \log p(x, z) \) is the log-likelihood of the complete observation $(x,z)$, and \( z \) is a latent variable. 
The M-step updates the current estimates with the arguments maximizing the $Q$-function: $p' \leftarrow \arg\max_p Q(p | p')$ . The sequence is repeated until convergence.

\subsection{Problem Scenario}
Let $x=(x_1, x_2, \ldots, x_K)$ denote $K$ inputs and the output be $y$. For simplicity, we limit $y$ to be continuous numerical, although this assumption could be relaxed to discrete or categorical without loss of generality. Component $x_k$ is a source: a modality, a single or group of variables, or an unstructured input such as an image or text. For clarity, we limit to $K=2$, where $x_1$ and $x_2$ are single variables.


Let $p(y|x)$ be the density function of the outcome given the inputs. A founding assumption, in the spirit of VAE, is that $p(y|x)$ decomposes into $p(y|x)=\int p(y|z) p(z_1|x_1) p(z_2|x_2)dz_1dz_2$, where $z=(z_1,z_2)$ are unobserved latent variables. We assume that $p(y|z,x)=p(y|z)$, that is, $z$ contains all the information of $x$ about $y$, and that $p(z|x)=p(z_1|x_1)p(z_2|x_2)$, that is, there is one latent variable per source. These are independent, conditionally on their corresponding source. While jointly embedding all features with a single model might seem more intuitive, our separate embedding approach offers several advantages: (1) it naturally handles multi-modal data where different feature groups may require distinct processing (e.g., images vs. tabular data), (2) it allows for modular model selection where each $g_{\phi_k}$ can be tailored to the characteristics of $x_k$, and (3) it can potentially lead to to handling missing data by allowing inference on available modalities. This design choice prioritizes flexibility, though the multi-modal setting and handling of missing data are left for future exploration.

The contribution to the log-likelihood of a complete observation $(y, z, x)$ is
$\log p(y , z| x) = \log p(y | z) + \log p(z| x)$. In the E-step, we compute
\begin{eqnarray}
\label{eqn:E_gen}
\int \log p(y , z| x) p'(z | y, x) dz &=& \int \log p(y | z) p'(z|y, x) dz + \int \log p(z| x) p'(z|y, x) dz.
\end{eqnarray}
\autoref{eqn:E_gen} can be estimated by MC sampling. Since sampling from $p'(z|y, x)$ can be inefficient, we rather rely on the decomposition $p'(z|y, x) = p'(y|z) p'(z|x)/p'(y|x)$. Thus, we sample $z_r$ from $p'(z|x)$, $r=1,\ldots,R$, and, setting $w_r=p'(y|z_r)/\sum_{t} p'(y|z_{t})$, approximate the right-hand side term of \autoref{eqn:E_gen}
\begin{eqnarray}
\label{eqn:E_MC_gen}
\int \log p(y , z| x) p'(z|y, x) dz &\approx& \sum_{r=1}^R \left\{\log p(y | z_r) + \log p(z_r| x)\right\} w_r.
\end{eqnarray}

\subsection{Objective Function}
\label{objective_function}
Adapting \autoref{eqn:E_MC_gen} for the observed data $\{(y_i,x_i)\}_{i=1}^N$, the overall loss function, $\mathcal{L}$, is
\begin{equation}
\label{eqn:overall_loss}
\mathcal{L}(\phi, \theta) = -\sum_{i=1}^N \sum_{r=1}^R\left\{\log p_{\phi}(z_{i,r} | x_i) + \log p_\theta(y_i | z_{i,r})\right\} w_{i,r},
\end{equation}
where the models of $p(y|z)$ and $p(z|x)$ inherit parameters \(\theta\) and \( \phi\), respectively. The weights are 
\begin{equation}
\label{eqn:compute_w}
w_{i,r} = \frac{p_{\theta'}(y_i | z_{i,r})}{\sum_{t=1}^R p_{\theta'}(y_i | z_{i,t})}
\end{equation}
where $z_i,r \stackrel{ind.}{\sim} p_{\phi'}(z|x_i)$, $r=1,\ldots,R$.

\autoref{eqn:overall_loss} shows that ${\cal L}$ is a sum of losses associated with the encoder model, $p_\phi$, for each source, and the decoder model, $p_\theta$, from the latent variables to the output. $p_{\theta}$ and $p_{\phi}$ are referred to as encoder and decoder in the spirit of auto-encoder models. At each M-step, ${\cal L}$ is minimized with respect to $\theta$ and $\phi$. Then, $\theta'$ and $\phi'$ are updated, as well as the weights and the sampled $z$. Then, the process is iterated until convergence.

\subsubsection{Example: \texorpdfstring{$\mathcal{L}(\phi, \theta)$}{} under normality}
For illustration purposes, we develop below the case where $p_\phi$ and $p_\theta$ are normal distributions, similar to \citet{kingma2014vae}. As a reminder, any other distributions could be adopted, including non-continuous or non-numerical outcomes. 


\paragraph{Encoder $p_{\phi}(z|x)$.} Let $m_k$ be the length of the latent variable $z_k$, $k=1,2$. We assume a normal model for $Z_k$ given $X_k=x_k$,
\begin{equation}
\label{eqn:compute_z}
Z_k | X_k=x_k \sim {\cal N}_{m_k}(g_{\phi_k}(x_k), \sigma_{k}^2 J_{m_k}),
\end{equation} 
where $J_{m_k}$ is the $m_k\times m_k$ identity matrix. In particular,
\begin{eqnarray}
\log p_\phi(z_k|x_k) &=& -\frac{m_k}{2}\log 2\pi - \frac{m_k}{2}\log \sigma_k^2 - \frac{1}{2\sigma_k^2} \sum_{j=1}^{m_k} \{z_{k,j} - g_{\phi_k,j}(x_k)\}^2, \quad k=1,2.
\end{eqnarray}
The mean $g_{\phi_k}(x_k)$ can be any model, such as a neural network, with output of length $m_k$, $k=1,2$. The scale $\sigma_{k}$ can be fixed, computed via the weighted residuals, or learned through a separate set of models. It controls the amount of noise introduced in the latent dimension and is pivotal in determining the width of the prediction interval for $p(y|z)$. The code implementation allows for all three methods. This paper only presents the results for a fixed $\sigma_{k}$, as well as training a separate set of $g_{\phi_k}(x_k)$ models to estimate varying $\sigma_{k}$s.


\paragraph{Decoder $p_{\theta}(y|z)$.}
We assume a normal model for $Y$ given $Z=z$,
\begin{equation}
\label{eqn:y_given_z}
    Y | Z=z \sim \mathcal{N}(f_\theta(z), \sigma^2).
\end{equation}
This results in a log-likelihood contribution,
\begin{equation}
\label{eqn:p_theta_llk}
    \log p_{\theta}(y|z) = -\frac{1}{2}\log 2\pi - \frac{1}{2}\log \sigma^2 - \frac{1}{2\sigma^2} \{y-f_\theta(z)\}^2.
\end{equation}
Again, the mean $f_\theta(z)$ can be any model, such as a neural network.



\paragraph{Summary.}
Overall, the M-step is

\begin{equation}
\phi_k^* = \arg\min_{\phi_k} \sum_{i,r} w_{i,r} \sum_{j=1}^{m_k} \{z_{k,i,r,j} - g_{\phi_k,j}(x_{k,i,r})\}^2,\quad k=1,2,
 \end{equation}
\begin{equation}
\theta^* = \arg\min_\theta \sum_{i,r} w_{i,r}\{y_i - f_{\theta}(z_{i,r})\}^2,
\end{equation}

\begin{equation}
\label{eq_sigma}
(\sigma^{*})^2 = \frac{1}{N} \sum_{i,r} w_{i,r}\{y_i - f_{\theta^*}(z_{i,r})\}^2,
\end{equation}



\subsection{Training}
\label{semf_training}
For efficiency purpose, the training set, $\{1, \ldots, N\}$, is segmented into batches \( \{b_1, \ldots, b_L\} \) on which the index $i$ runs (and thus the denominator of \autoref{eq_sigma} must be adapted accordingly). The process iterates for each batch until the maximum number of steps is reached or an early stopping criterion is satisfied. The full details are given in \autoref{algo:semf-train}. The framework requires tuning hyper-parameters such as the number of MC samples $R$, the number of latent nodes $m_k$, and the standard deviation $\sigma_{k}$ of $Z_k$. Monitoring the point prediction on a hold-out validation is important to combat overfitting and terminate the training early with a \textsc{patience} hyper-parameter. Moreover, due to the generative nature of SEMF, the variation resulting from the initial random seed is measured in \autoref{experiment_models}. Additionally, the model-specific hyper-parameters ($p_\phi$ and $p_\theta$) are also discussed in the same subsection.

\begin{algorithm}[H]
\caption{SEMF Training: two input sources}
\label{algo:semf-train}
\begin{algorithmic}[1]
\REQUIRE \( y, x_1, x_2, R \) \COMMENT{Training data and number of MC samples}
\ENSURE \( \theta, \phi_1, \phi_2 \) \COMMENT{Trained model parameters}
\STATE Initialize \( \theta, \phi_1, \phi_2 \)
\STATE Initialize \( D_y, D_{z_1}, D_{z_2} \) to \( \emptyset \) \COMMENT{Data buffers for batch updates}
\STATE Split \( I = \{1, \ldots, N\} \) into \( L \) batches \( \{b_1, \ldots, b_L\} \)

\FOR{\( \ell = 1 \), \ldots, \( L \)}
    \FORALL{\( i \) in \( b_\ell \)}
            \FOR{\( r = 1 \), \ldots, \( R \)}
                \STATE Simulate \( z_{1,i,r} \sim p_{\phi_1}(\cdot | x_{1,i}) \)
                \STATE Simulate \( z_{2,i,r} \sim p_{\phi_2}(\cdot | x_{2,i}) \)
                \STATE Set \( z_{i,r} = [z_{1,i,r}, z_{2,i,r}] \)
            \ENDFOR
        \FOR{\( r = 1 \), \ldots, \( R \)}
            \STATE Compute 
            \[ w_{i,r} = \frac{p_{\theta}(y_i | z_{i,r})}{\sum_{t=1}^R p_{\theta}(y_i | z_{i,t})} \]
            \STATE Update \( D_y \leftarrow D_y \cup [y_i | z_{i,r} | w_{i,r}] \)
            \STATE Update \( D_{z_1} \leftarrow D_{z_1} \cup [z_{1,i,r} | x_{1,i} | w_{i,r}] \)
            \STATE Update \( D_{z_2} \leftarrow D_{z_2} \cup [z_{2,i,r} | x_{2,i} | w_{i,r}] \)
        \ENDFOR
    \ENDFOR
    \STATE Update \( \theta \leftarrow \arg\min_\theta \sum_{(y,z,w) \in D_y} w \cdot \mathcal{L}_\theta(y, z) \) \COMMENT{M-step for decoder}
    \STATE Update \( \phi_1 \leftarrow \arg\min_{\phi_1} \sum_{(z,x,w) \in D_{z_1}} w \cdot \mathcal{L}_{\phi_1}(z, x) \) \COMMENT{M-step for encoder 1}
    \STATE Update \( \phi_2 \leftarrow \arg\min_{\phi_2} \sum_{(z,x,w) \in D_{z_2}} w \cdot \mathcal{L}_{\phi_2}(z, x) \) \COMMENT{M-step for encoder 2}
    \STATE Clear \( D_y, D_{z_1}, D_{z_2} \) \COMMENT{Reset buffers for next batch}
\ENDFOR
\STATE Check convergence; Go to step 4 if not
\end{algorithmic}
\end{algorithm}

\subsection{Inference}
\begin{figure}[ht]
    \begin{center}
    \includegraphics[width=0.8\columnwidth]{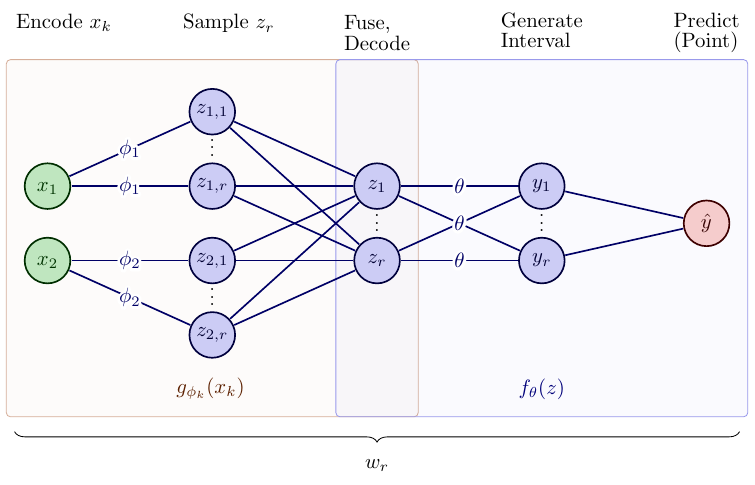}
    \caption{Inference procedure with the SEMF's learnable parameters $\phi_k$ and $\theta$. Here, we illustrate the number of input sources $k$ as $k=1,2$
    }
    \label{fig:semf-overview}
    \end{center}
\end{figure}

The encoder-decoder structure of SEMF entails the simulations of $z_{r}$ during inference, as depicted in \autoref{fig:semf-overview}. In theory, any inference on $y$ given $x$ can be performed for $\hat{y}$, for instance the mean value $\hat{y} = \frac{1}{R} \sum_{r=1}^R f_\theta(z_r)$, where $z_r \sim p_\phi(z|x)$ (see \autoref{algo:semf-infer} for the simulation scheme). For prediction intervals, a second simulation step is used,

\begin{equation}
  z_r \sim p(z|x), \quad \hat{y}_{r,s} \sim p_{\theta}(y | z_r),\quad r,s=1,\ldots,R.
\end{equation}
Prediction interval on $y$ given $x$ at a given level $\alpha$ follows as
\begin{equation}
\label{semf_quantile}
PI = \text{quantile}\left(\{\hat{y}_{r,s}\}; \frac{\alpha}2, 1-\frac{\alpha}2\right).
\end{equation}
In order to calibrate these intervals, we can use conformal prediction \citep{vovk2005algorithmic, romano2019cqr}, which we review in \autoref{related_work:prediction-intervals} and subsequently explain in \autoref{experiment_models}. We primarily use Conformalized Quantile Regression (CQR) from \citet{romano2019cqr} as our calibration method. While SEMF generates intervals through latent variable sampling, we apply CQR to these raw intervals, ensuring valid marginal coverage while assessing the underlying interval generation quality of SEMF. The CQR procedure is detailed in \autoref{algo:cqr} below, which extends the general split conformal framework for quantile regression methods. CQR first trains separate quantile regressors to estimate the $\alpha/2$ and $1-\alpha/2$ quantiles, then uses a calibration set to compute conformity scores as $S_i = \max\{\hat{q}_{\alpha/2}(X_i) - Y_i, Y_i - \hat{q}_{1-\alpha/2}(X_i)\}$. The method then adjusts the initial quantile predictions by adding a correction factor derived from the empirical quantile of these conformity scores, ensuring finite-sample coverage guarantees. CQR has proven particularly effective for regression problems where the underlying model can produce reasonable quantile estimates, making it a strong baseline for uncertainty quantification tasks.


\begin{algorithm}
    \caption{SEMF Inference: single test example}
    \label{algo:semf-infer}
    \begin{algorithmic}[1]
    \REQUIRE $\theta^*, \phi_1^*, \phi_2^*, x_1, x_2, R$ \COMMENT{Trained parameters and test input}
\ENSURE $\{z_r\}_{r=1}^R$ \COMMENT{R samples from the latent distribution}
\FOR{$r = 1, \ldots, R$}
        \STATE Simulate $z_{1,r} \sim p_{\phi_1^*}(\cdot | x_1)$ \COMMENT{Sample from encoder 1}
        \STATE Simulate $z_{2,r} \sim p_{\phi_2^*}(\cdot | x_2)$ \COMMENT{Sample from encoder 2}
        \STATE Set $z_r = [z_{1,r}, z_{2,r}]$ \COMMENT{Concatenate latent representations}
\ENDFOR
\STATE \COMMENT{Applications in Section 2.4}
\STATE \textbf{Point prediction:} $\hat{y} \leftarrow \frac{1}{R} \sum_{r=1}^R f_{\theta^*}(z_r)$
\STATE \textbf{Prediction intervals:} $\hat{y}_{r,s} \sim p_{\theta^*}(y | z_r)$ for $s=1,\ldots,R$
    \end{algorithmic}
\end{algorithm}

\begin{algorithm}[H]
\caption{Conformalized Quantile Regression (CQR)}
\label{algo:cqr}
\begin{algorithmic}[1]
\REQUIRE Training data $\mathcal{D}_{train}$, calibration data $\mathcal{D}_{calib} = \{(X_i, Y_i)\}_{i=1}^{n_{calib}}$, test input $X_{test}$, significance level $\alpha$
 \ENSURE Prediction interval $[L_{CQR}, U_{CQR}]$ with coverage guarantee $\mathbb{P}(Y_{test} \in [L_{CQR}, U_{CQR}]) \geq 1-\alpha$

\STATE Train quantile regressors $\hat{f}^{\alpha/2}$ and $\hat{f}^{1-\alpha/2}$ on $\mathcal{D}_{train}$
\STATE \COMMENT{Step 1: Compute initial quantile predictions on calibration set}
\FOR{$i = 1, \ldots, n_{calib}$}
    \STATE $\hat{L}_i \leftarrow \hat{f}^{\alpha/2}(X_i)$ \COMMENT{Lower quantile prediction}
    \STATE $\hat{U}_i \leftarrow \hat{f}^{1-\alpha/2}(X_i)$ \COMMENT{Upper quantile prediction}
\ENDFOR

\STATE \COMMENT{Step 2: Compute conformity scores on calibration set} 
\FOR{$i = 1, \ldots, n_{calib}$}
    \STATE $S_i \leftarrow \max\{\hat{L}_i - Y_i, Y_i - \hat{U}_i\}$ \COMMENT{CQR conformity score}
\ENDFOR

\STATE \COMMENT{Step 3: Find quantile correction factor}
\STATE Sort $\{S_i\}_{i=1}^{n_{calib}}$ in increasing order
\STATE $\hat{q} \leftarrow \lceil(1-\alpha)(n_{calib}+1)\rceil$-th smallest score \COMMENT{If index $>$ $n_{calib}$, use $\max_i S_i$}

\STATE \COMMENT{Step 4: Apply correction to test prediction} 
\STATE $\hat{L}_{test} \leftarrow \hat{f}^{\alpha/2}(X_{test})$, $\hat{U}_{test} \leftarrow \hat{f}^{1-\alpha/2}(X_{test})$
\STATE $L_{CQR} \leftarrow \hat{L}_{test} - \hat{q}$, $U_{CQR} \leftarrow \hat{U}_{test} + \hat{q}$

\STATE \textbf{return} $[L_{CQR}, U_{CQR}]$
\end{algorithmic}
\end{algorithm}

\section{Related Work}
\label{related_work}

\subsection{Latent Representation Learning}
Latent-representation learning typically relies on the encoder-decoder architecture introduced in \autoref{method}.  Classic examples are Auto-Encoders (AEs) and VAEs: the encoder \(g_{\phi}\) maps a sample input \(x\) to a latent variable \(z\), and the decoder \(f_{\theta}\) reconstructs the input \(\hat{x}=f_{\theta}(z)\). Training jointly optimizes the parameter vectors \(\phi\) and \(\theta\) by minimizing the reconstruction loss. Unlike AEs, which focus solely on reconstruction, VAEs introduce a variational objective that aims to model the data distribution through the marginal likelihood $p_\theta(x)$ while fitting approximate posterior distributions over the latent variables \citep{kingma2014vae,blei2017vireview}. However, directly maximizing this likelihood is difficult \citep{blei2017vireview}. Thus, alternative methods exist, such as maximizing the ELBO, which provides guarantees on the log-likelihood \citep{balakrishnan2017emguarantees}.

For supervised and semi-supervised tasks, latent representation learning can include task-specific predictions \citep{kingma2014semi}. More specifically, models such as AEs follow the classical encoder-decoder objective while training a predictor $h_{\psi}(z)$ through an additional layer or model to estimate the output $y$. This dual objective facilitates the learning of more task-relevant embeddings \citep{zhuang2015supervisedautoencoder, le2018supervisedae}. Semi-supervised VAEs are similar, with the distinction that they couple the reconstruction loss of the unlabeled data with a variational approximation of latent variables. This is effective even with sparse labels \citep{ji2020Multisupervisedvae, zhuang2023semi}. 

The EM algorithm has already been used for supervised learning tasks using specific models \citep{ghahramani1993em, williams2005, louiset2021ucsl}, where the goal has been point prediction with GMMs. Similarly, the EM algorithm adapts well to minimal supervision \citep{luo2020weakly}, as well as using labeled and unlabeled data in semi-supervised settings for both single and multiple modalities \citep{he2022semisupem, xu2024expectation}. Our work differs in that we modify the use of MC sampling to generate prediction intervals with any ML model, and in theory, under any distribution.

\subsection{Prediction Intervals}
\label{related_work:prediction-intervals}

Crucial for estimating uncertainty, prediction intervals provide a range for regression outcomes. Common approaches include Bayesian methods \citep{williams1995gaussian,hensman2015scalable,gal2016dropoutbnn}, ensemble techniques \citep{breiman2001random,lakshminarayanan2017simple,malinin2021uncertainty}, and quantile regression \citep{Koenker1978,koenker2001qr}. Quantile regression specifically utilizes the pinball loss function to target desired quantiles, proving effective even for non-parametric models \citep{steinwart2011estimating} and asymmetric distributions \citep{koenker2001qr}. This loss is valuable both for individual models and for refining quantile estimates when used with ensembles \citep{meinshausen2006quantile}—a loss that can also be learned jointly as demonstrated by Simultaneous Quantile Regression (SQR) \citep{tagasovska2019single}. SQR represents a powerful neural network approach that jointly estimates multiple quantiles while maintaining their proper ordering, making it particularly effective for constructing prediction intervals and a strong baseline for uncertainty quantification tasks.

Complementary to these approaches, conformal prediction (CP) provides a general framework for constructing and calibrating prediction intervals for any model \citep{vovk2022conformal}. Assuming data exchangeability, CP adjusts intervals to ensure they meet a pre-specified coverage probability, thereby enhancing reliability in applications demanding rigorous uncertainty quantification. Recent advances in CP include its integration with ensemble models for time series \citep{xu21h, xu23r} and Bayesian approaches, such as Conformal Bayesian model averaging (CBMA), which aggregates conformity scores from several Bayesian models to obtain optimal prediction sets with guarantees \citep{bhagwat2025cbma}. Furthermore, several CP frameworks that incorporate data-dependent score functions have recently been proposed to improve the efficiency and validity of intervals derived via quantile estimators \citep{ge2024optimal}, boosting methods (XGBoost) to refine conformity scores \citep{xie2024boosted}, and locally adaptive techniques to enhance conditional validity \citep{colombo2023on}.

\section{Experimental Setup}
\label{experimental_setup}


\subsection{Models}
\label{experiment_models}
Our baseline consists of quantile regression models based on eXtreme Gradient Boosting (XGBoost) \citep{Chen2016}, Extremely Randomized Trees (ET) \citep{geurts2006extremely},
and neural networks using SQR \citep{tagasovska2019single}, all summarized and depicted in \autoref{tab:model_comparisons}. To ensure consistency in our experimental setup, we align the families and hyper-parameters of $p_{\phi}$ and $p_{\theta}$ with our baseline models. For example, in the case of XGBoost in SEMF, we use $K$ XGBoosts, $g_{\phi_k}(x_k)$, one for each input $x_k$, $k=1,\ldots, K$, and one XGBoost for $f_\theta(z)$ with the same hyper-parameters. We refer to the SEMF's adoption of these models as MultiXGBs, MultiETs, and MultiMLPs. When establishing prediction intervals, we conformalize our prediction intervals according to \citep{romano2019cqr} at an uncertainty tolerance of 5\% for both the baseline and SEMF (\autoref{semf_quantile}). 

Regarding notable hyperparameters, we target larger $\sigma_k$ values that introduce more noise and produce better intervals than point predictions. However, if the primary goal is to produce better point predictions, this hyperparameter should be set to a smaller value. We do not present the point prediction results in this paper, though they can be found in our code implementation. For several datasets, we set $\sigma_{k}$ to \texttt{train\_residual\_models}, which trains separate models to predict the scale of the latent variables individually and is more suitable for heteroscedastic noise patterns. This approach often yields substantially improved interval predictions by capturing local uncertainty structures, though at the cost of longer computation times due to the additional model training. In our experience, using \texttt{train\_residual\_models} for $\sigma_{k}$ is recommended as a starting point for new applications, with fixed values being a computationally efficient alternative once the behavior of the dataset is better understood. The optimal models are then trained and tested with five different seeds, and the results are averaged, and the variability of the results is presented. The point prediction, $\hat{y}$, uses the mean inferred values. \autoref{appendix:optimal-semf} contains more details on the hyper-parameters for each SEMF model and dataset introduced below.

\begin{table}[!htbp]
\centering
\begin{small}
\resizebox{\textwidth}{!}
{\begin{tabular}{p{2cm} p{6cm} p{6cm}}
\toprule
\textbf{SEMF} & \multicolumn{1}{c}{\textbf{Base Model}} & \multicolumn{1}{c}{\textbf{Interval Prediction Baseline}} \\
\midrule
MultiXGBs & XGBoost \newline \small{Trees: 100, Maximum depth: 6, Early stopping steps: 10} & Quantile XGBoost \newline \small{Same as point prediction baseline, XGBoost} \\
\cmidrule(lr){1-3}
MultiETs & Extremely Randomized Trees \newline \small{Trees: 100, Maximum depth: 10} & Quantile Extremely Randomized Trees \newline \small{Same as point prediction baseline, Extremely Randomized Trees} \\
\cmidrule(lr){1-3}
MultiMLPs & Deep Neural Network \newline \small{Hidden layers: 2, Nodes per layer: 100, Activation functions: ReLU, Epochs: 1000 or 5000, Learning rate: 0.001, Batch training, Early stopping steps: 100} & Simultaneous Quantile Regression \newline \small{Same as point prediction baseline, Deep Neural Network} \\
\bottomrule
\end{tabular}
}
\end{small}
\caption{SEMF models, baselines, and hyper-parameters.}
\label{tab:model_comparisons}
\end{table}

\subsection{Datasets}
\subsubsection{Simulations}

\begin{figure}[!h]
    \begin{center}
    \includegraphics[width=0.9\columnwidth]{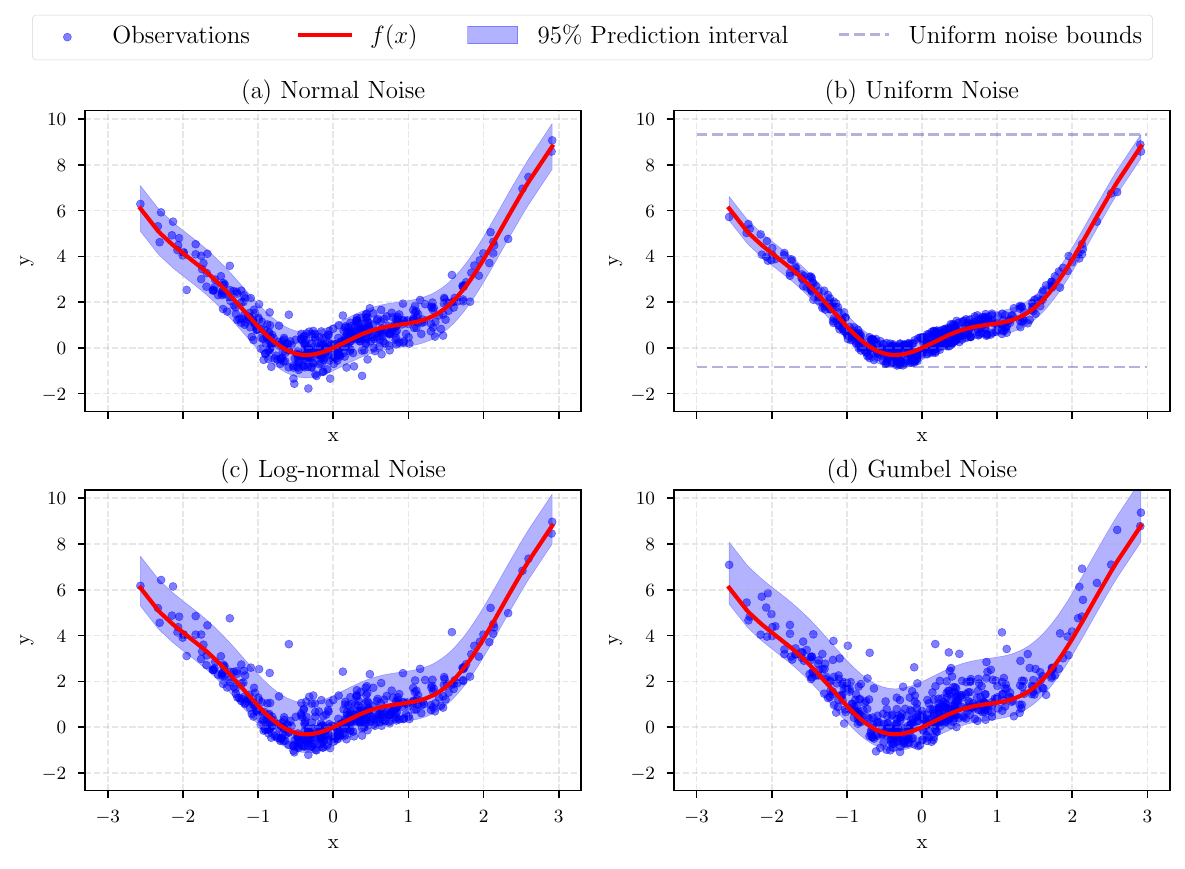}
    \caption{Prediction intervals under different noise distributions for a one-dimensional $x$ generated from 500 observations. Each panel shows predictions (blue dots) with 95\% prediction intervals (shaded regions) for a model trained on data with (a) Normal, (b) Uniform, (c) Log-normal, and (d) Gumbel noise. The red curve denotes \(f(x)\), the underlying (deterministic) function.}
    \label{fig:synthetic_prediction_intervals}
    \end{center}
\end{figure}

In our experiments, we generate synthetic datasets with \(10^3\) observations and $k=2$ predictors from a standard normal distribution according to 
\begin{equation}
  y = f(x) + \epsilon,
\end{equation}
where \(f(x)\) in a simple setup is defined as 
\begin{equation}
  f(x)=\sum_{i=1}^{k}\cos(x_i)
\end{equation}
to isolate noise effects, and in a more complex setup, we instead generate
\begin{equation}
  \label{eqn:quadratic_plus_periodic}
  f(x)=\sum_{i=1}^{k}\Bigl(x_i^2+0.5\,\sin(3\,x_i)\Bigr)
\end{equation}
to introduce nonlinearity and heteroscedasticity. \(\epsilon\) is drawn from one of four distributions: \(\mathcal{N}(0,0.5)\), \(\mathcal{U}(-0.5,0.5)\), a centered log-normal, or \(\text{Gumbel}(0,0.5)\). \autoref{fig:synthetic_prediction_intervals} illustrates an example of \autoref{eqn:quadratic_plus_periodic} with $k=1$ (for simplicity) and 500 observations, where blue dots indicate predictions with 95\% prediction intervals and the red curve denotes \(f(x)\).

For our simulation experiments, the underlying data is generated from a fixed seed with variability only in the model seeds (\autoref{experiment_models}). We evaluate all three model variants (MultiXGBs, MultiETs, and MultiMLPs) against standard quantile regression baselines. For MultiXGBs and MultiETs, we use $R=10$ with 10 nodes per latent dimension, while MultiMLPs use $R=5$ with 20 nodes per latent dimension. All models employ early stopping with appropriate patience settings to prevent overfitting on the synthetic data. The data scaling and splits are the same as the benchmark data below.

\subsubsection{Benchmark data}
We systematically curate a subset of datasets from the OpenML-CTR23 \citep{fischer2023}  benchmark suite to evaluate and carry out our experiments. Initially comprising 35 datasets, we apply an exclusion criteria to refine this collection to 11 datasets. The details and overview are in \autoref{appendix:semf-databench}. We remove duplicated rows from all the datasets and carry out the scaling of all predictors, including the outcome. The features of these datasets are then treated as separate inputs to SEMF.
In all our datasets, 70\% of the data is used to train all models, 15\% as a hold-out validation set to monitor SEMF's performance, and 15\% to evaluate the models. To combat overfitting, baseline models that benefit from early stopping are allocated another 15\% from the training data. Lastly, it is essential to note that all data in SEMF are processed batch-wise, without employing mini-batch training, to ensure consistency and stability in the training process.

\subsection{Metrics}

The evaluation of prediction intervals employs Prediction Interval Coverage Probability (PICP), Normalized Mean Prediction Interval Width (NMPIW), Continuous Ranked Probability Score (CRPS) and the quantile (pinball) loss. Detailed definitions and closed-form expressions for these metrics (with the CRPS computed under a uniformity assumption on the predictive distribution) are provided in \autoref{appendix:metrics}. In addition, we introduce the Coverage-Width Ratio (CWR)
\begin{equation}
    \text{CWR}=\frac{\text{PICP}}{\text{NMPIW}},
\end{equation}
which quantifies the trade-off between reliability and precision. 

In our case, measuring the performance of SEMF over the baseline models is far more critical than reviewing absolute metrics in isolation. For any metric above, this is computed as 
\small
\begin{equation}
\label{eqn:base_metric}
\text{Metric}_\Delta(\%) = \left( \frac{\text{Metric}_{\text{SEMF}} - \text{Metric}_{\text{Baseline}}}{\text{Metric}_{\text{Baseline}}} \right) \times 100 ,
\end{equation}
\normalsize
on which we base our decisions for selecting the best hyper-parameters as explained in \autoref{appendix:relative-metrics-impact}.

\section{Experiments}
\label{results}

\subsection{Simulated Data}
\label{simulated-data}

\autoref{tab:synthetic-cosine-x2-results} and \autoref{tab:synthetic-quadratic-x2-results} summarize the performance of our SEMF variants on synthetic datasets generated using two predictors. The PICP shows that on average, all models across all experiments were able to achieve at least the 95\% desired coverage probability.

\begin{table}[!htbp]
\centering%
\vskip 0.15in
\begin{small}

\resizebox{0.8\columnwidth}{!}{%
\begin{tabular}{lcccccc}
\toprule
 & $\Delta$CWR & $\Delta$NMPIW & $\Delta$CRPS & $\Delta$Pinball & $\Delta$PICP & PICP \\
\midrule
\multicolumn{7}{l}{\textbf{MultiXGBs}} \\
normal & \textbf{18\% [10:35]} & \textbf{16\% [10:28]} & \textbf{6\% [3:10]} & \textbf{11\% [5:17]} & -1\% [-3:0] & 0.95$\pm$0.01 \\
uniform & \textbf{25\% [-2:58]} & \textbf{19\% [-5:39]} & \textbf{10\% [-1:23]} & \textbf{18\% [0:32]} & -1\% [-5:2] & 0.96$\pm$0.03 \\
lognormal & \textbf{28\% [8:59]} & \textbf{21\% [8:39]} & \textbf{13\% [8:18]} & \textbf{7\% [2:14]} & -1\% [-3:0] & 0.96$\pm$0.01 \\
gumbel & \textbf{10\% [-5:20]} & \textbf{8\% [-9:18]} & \textbf{8\% [1:14]} & \textbf{8\% [0:16]} & 0\% [-3:4] & 0.96$\pm$0.01 \\
\multicolumn{7}{l}{\textbf{MultiETs}} \\
normal & \textbf{15\% [1:32]} & \textbf{14\% [0:27]} & \textbf{5\% [0:8]} & \textbf{8\% [0:16]} & -2\% [-4:1] & 0.95$\pm$0.01 \\
uniform & \textbf{16\% [3:35]} & \textbf{13\% [-1:28]} & \textbf{6\% [3:10]} & \textbf{13\% [8:21]} & 0\% [-3:4] & 0.97$\pm$0.01 \\
lognormal & \textbf{5\% [-14:18]} & \textbf{3\% [-18:17]} & \textbf{9\% [5:13]} & -3\% [-10:2] & 0\% [-2:2] & 0.97$\pm$0.01 \\
gumbel & \textbf{3\% [-4:14]} & \textbf{3\% [-7:14]} & \textbf{8\% [6:11]} & \textbf{1\% [-2:5]} & 0\% [-2:2] & 0.96$\pm$0.01 \\
\multicolumn{7}{l}{\textbf{MultiMLPs}} \\
normal & \textbf{3\% [-2:9]} & \textbf{3\% [-2:8]} & \textbf{1\% [-2:4]} & \textbf{2\% [-3:5]} & 0\% [-1:0] & 0.96$\pm$0.01 \\
uniform & \textbf{4\% [-2:13]} & \textbf{4\% [-2:10]} & \textbf{2\% [-1:8]} & \textbf{7\% [-5:23]} & 0\% [-4:3] & 0.97$\pm$0.02 \\
lognormal & \textbf{7\% [-4:18]} & \textbf{6\% [-5:16]} & \textbf{3\% [-2:7]} & \textbf{4\% [0:9]} & 0\% [-1:1] & 0.97$\pm$0.01 \\
gumbel & \textbf{5\% [-3:13]} & \textbf{5\% [-5:14]} & 0\% [-2:3] & 0\% [-4:4] & -1\% [-2:2] & 0.96$\pm$0.02 \\
\bottomrule
\end{tabular}
}

\end{small}
\caption{Test results (95\% prediction intervals) for 1000 observations generated using cosine $f(x)$ with 2 predictors, and an additive noise ($\epsilon$) belonging to one of the four distributions. Relative metrics are shown over 5 seeds as `mean [min:max]', and absolute metrics as `mean $\pm$ std'. Average performance over the baseline is highlighted in bold.}
\label{tab:synthetic-cosine-x2-results}
\end{table}

\begin{table}[!htbp]
\centering%
\vskip 0.15in
\begin{small}

\resizebox{0.8\columnwidth}{!}{%
\begin{tabular}{lcccccc}
\toprule
 & $\Delta$CWR & $\Delta$NMPIW & $\Delta$CRPS & $\Delta$Pinball & $\Delta$PICP & PICP \\
\midrule
\multicolumn{7}{l}{\textbf{MultiXGBs}} \\
normal & \textbf{39\% [28:65]} & \textbf{28\% [22:41]} & \textbf{29\% [26:34]} & \textbf{23\% [16:35]} & 0\% [-3:0] & 0.95$\pm$0.01 \\
uniform & \textbf{65\% [54:80]} & \textbf{39\% [34:44]} & \textbf{41\% [39:47]} & \textbf{25\% [16:32]} & 0\% [-1:1] & 0.96$\pm$0.02 \\
lognormal & 0\% [-13:7] & -2\% [-16:6] & \textbf{20\% [14:27]} & \textbf{4\% [-4:16]} & \textbf{2\% [0:5]} & 0.96$\pm$0.02 \\
gumbel & \textbf{23\% [0:49]} & \textbf{17\% [0:35]} & \textbf{20\% [9:30]} & \textbf{16\% [3:21]} & 0\% [-3:4] & 0.96$\pm$0.01 \\
\multicolumn{7}{l}{\textbf{MultiETs}} \\
normal & \textbf{23\% [7:35]} & \textbf{19\% [7:26]} & \textbf{14\% [0:21]} & \textbf{12\% [-4:23]} & -1\% [-2:1] & 0.95$\pm$0.02 \\
uniform & \textbf{40\% [25:71]} & \textbf{28\% [20:42]} & \textbf{24\% [16:33]} & \textbf{12\% [-4:21]} & 0\% [-3:2] & 0.97$\pm$0.02 \\
lognormal & \textbf{8\% [-9:33]} & \textbf{5\% [-13:27]} & \textbf{11\% [0:16]} & \textbf{6\% [-6:12]} & \textbf{1\% [-2:3]} & 0.97$\pm$0.01 \\
gumbel & \textbf{29\% [-3:68]} & \textbf{19\% [-4:42]} & \textbf{18\% [5:30]} & \textbf{15\% [0:28]} & 0\% [-2:1] & 0.97$\pm$0.01 \\
\multicolumn{7}{l}{\textbf{MultiMLPs}} \\
normal & \textbf{5\% [-14:24]} & \textbf{1\% [-20:21]} & \textbf{7\% [2:13]} & \textbf{13\% [6:20]} & \textbf{2\% [-2:4]} & 0.95$\pm$0.01 \\
uniform & \textbf{37\% [15:66]} & \textbf{25\% [13:41]} & \textbf{19\% [13:27]} & \textbf{29\% [9:40]} & \textbf{1\% [-2:6]} & 0.97$\pm$0.02 \\
lognormal & \textbf{1\% [-12:10]} & \textbf{1\% [-15:11]} & \textbf{4\% [2:8]} & -1\% [-10:4] & 0\% [-2:1] & 0.96$\pm$0.01 \\
gumbel & \textbf{15\% [-6:38]} & \textbf{11\% [-8:30]} & \textbf{10\% [-3:23]} & \textbf{11\% [-5:23]} & 0\% [-3:2] & 0.96$\pm$0.00 \\
\bottomrule
\end{tabular}
}

\end{small}
\caption{Test results (95\% prediction intervals) for 1000 observations generated using quadratic $f(x)$ with 2 predictors, and an additive noise ($\epsilon$) belonging to one of the four distributions. Relative metrics are shown over 5 seeds as `mean [min:max]', and absolute metrics as `mean $\pm$ std'. Average performance over the baseline is highlighted in bold.}
\label{tab:synthetic-quadratic-x2-results}
\end{table}

For the cosine experiments (\autoref{tab:synthetic-cosine-x2-results}), all three methods (MultiXGBs, MultiETs, and MultiMLPs) exhibit positive relative improvements over the baseline in terms of the CWR, NMPIW, CRPS, and pinball loss. For instance, under normal noise, the MultiXGBs variant improves the $\Delta$CWR by an average of 18\% (ranging between 10\% and 35\%), while still preserving a PICP of approximately 0.95. Similar positive trends are seen across the uniform and lognormal noise conditions. Although under Gumbel noise the improvements are less pronounced, SEMF still delivers performance that is at least on par with the baseline.

For the quadratic experiments (\autoref{tab:synthetic-quadratic-x2-results}), which represents a more complex heteroscedastic case, the improvements are even more substantial. Under normal noise, MultiXGBs achieve an average improvement of 39\% in $\Delta$CWR, and, when the noise is uniform, gains can be as high as 65\%. There is some variability—for example, under lognormal noise, the improvement in $\Delta$CWR for MultiXGBs is close to zero—but overall, the results indicate that SEMF produces tighter and more reliable prediction intervals compared to the baseline. Similar to the cosine case, the Gumbel noise setting shows the smallest relative gains; nevertheless, the performance under Gumbel noise remains competitive with the baseline.

\subsection{Benchmark Data}

We trained and tested 165 models corresponding to the three model types—MultiXGBs, MultiETs, and MultiMLPs—across 11 datasets, using five seeds for each combination. \autoref{table-combined-complete-aggregated} presents the means and standard deviations for our metrics aggregated over the five seeds. \autoref{appendix:full-results} includes the results from each individual run.

\begin{table}[!htbp]
\centering%
\vskip 0.15in
\begin{footnotesize}
\setlength{\tabcolsep}{3.5pt}
\resizebox{\columnwidth}{!}{%
\begin{tabular}{lcccccccc}
\toprule

 & \multicolumn{5}{c}{Relative Metrics} & \multicolumn{3}{c}{Absolute Metrics} \\
\cmidrule(lr){2-6} \cmidrule(lr){7-9}
Dataset & $\Delta$CWR & $\Delta$NMPIW & $\Delta$CRPS & $\Delta$Pinball & $\Delta$PICP & CWR & NMPIW & PICP \\
\midrule
\multicolumn{9}{l}{\textbf{MultiXGBs}} \\ \cmidrule(lr){1-9}

space\_ga & \textbf{11\% [4:21]} & \textbf{9\% [2:19]} & \textbf{11\% [10:12]} & \textbf{9\% [6:13]} & 0\% [-2:2] & 4.51$\pm$1.64 & 0.24$\pm$0.08 & 0.95$\pm$0.01 \\
cpu\_activity & \textbf{5\% [-3:11]} & \textbf{5\% [-4:11]} & \textbf{16\% [10:26]} & \textbf{15\% [0:36]} & 0\% [-1:1] & 9.77$\pm$0.55 & 0.10$\pm$0.01 & 0.95$\pm$0.01 \\
naval\_propulsion\_plant & \textbf{163\% [114:200]} & \textbf{61\% [53:68]} & \textbf{72\% [66:74]} & \textbf{45\% [31:50]} & 0\% [-3:2] & 8.27$\pm$0.68 & 0.12$\pm$0.01 & 0.95$\pm$0.01 \\
miami\_housing & \textbf{5\% [-4:12]} & \textbf{5\% [-4:10]} & \textbf{12\% [8:17]} & -12\% [-21:-5] & 0\% [-1:0] & 8.75$\pm$0.36 & 0.11$\pm$0.00 & 0.95$\pm$0.00 \\
kin8nm & \textbf{18\% [16:19]} & \textbf{16\% [14:17]} & \textbf{23\% [20:25]} & \textbf{8\% [3:11]} & -1\% [-1:-1] & 2.31$\pm$0.04 & 0.41$\pm$0.01 & 0.94$\pm$0.00 \\
concrete\_compressive\_strength & \textbf{40\% [24:73]} & \textbf{29\% [20:45]} & \textbf{32\% [27:35]} & \textbf{12\% [4:25]} & -3\% [-5:-1] & 3.22$\pm$0.21 & 0.29$\pm$0.02 & 0.94$\pm$0.02 \\
cars & \textbf{32\% [10:67]} & \textbf{24\% [10:42]} & \textbf{27\% [20:42]} & \textbf{17\% [6:32]} & -1\% [-3:0] & 5.33$\pm$0.55 & 0.18$\pm$0.02 & 0.95$\pm$0.01 \\
energy\_efficiency & \textbf{128\% [72:230]} & \textbf{53\% [39:68]} & \textbf{67\% [59:81]} & \textbf{54\% [43:73]} & \textbf{3\% [-2:8]} & 13.95$\pm$2.46 & 0.07$\pm$0.01 & 0.96$\pm$0.00 \\
california\_housing & \textbf{1\% [-1:4]} & \textbf{1\% [-1:4]} & \textbf{18\% [15:23]} & -8\% [-10:-3] & 0\% [-1:0] & 2.24$\pm$0.08 & 0.42$\pm$0.02 & 0.95$\pm$0.01 \\
airfoil\_self\_noise & -8\% [-32:6] & -12\% [-47:6] & \textbf{2\% [-36:20]} & -17\% [-43:4] & 0\% [-1:0] & 2.29$\pm$0.38 & 0.44$\pm$0.10 & 0.97$\pm$0.01 \\
QSAR\_fish\_toxicity & \textbf{19\% [2:53]} & \textbf{14\% [1:35]} & \textbf{13\% [8:24]} & \textbf{15\% [8:30]} & 0\% [-1:2] & 1.71$\pm$0.10 & 0.57$\pm$0.04 & 0.98$\pm$0.01 \\
\cmidrule(lr){1-9}
\multicolumn{9}{l}{\textbf{MultiETs}} \\ \cmidrule(lr){1-9}

space\_ga & \textbf{5\% [-2:15]} & \textbf{6\% [-2:16]} & \textbf{8\% [6:11]} & -6\% [-9:-3] & -1\% [-3:0] & 4.38$\pm$1.82 & 0.26$\pm$0.09 & 0.95$\pm$0.02 \\
cpu\_activity & \textbf{2\% [-1:5]} & \textbf{6\% [2:10]} & \textbf{3\% [-2:7]} & -18\% [-22:-11] & -4\% [-5:-3] & 8.25$\pm$0.23 & 0.11$\pm$0.00 & 0.95$\pm$0.01 \\
naval\_propulsion\_plant & \textbf{144\% [118:160]} & \textbf{60\% [56:63]} & \textbf{68\% [64:72]} & \textbf{44\% [40:52]} & -2\% [-6:0] & 3.65$\pm$0.25 & 0.26$\pm$0.02 & 0.95$\pm$0.01 \\
miami\_housing & -9\% [-15:-5] & -10\% [-19:-4] & -3\% [-7:0] & -73\% [-81:-67] & 0\% [0:1] & 6.44$\pm$0.27 & 0.15$\pm$0.01 & 0.95$\pm$0.00 \\
kin8nm & \textbf{7\% [5:10]} & \textbf{7\% [4:8]} & \textbf{14\% [12:17]} & -1\% [-3:0] & 0\% [0:1] & 2.13$\pm$0.05 & 0.45$\pm$0.01 & 0.95$\pm$0.01 \\
concrete\_compressive\_strength & \textbf{6\% [-14:26]} & \textbf{4\% [-20:23]} & \textbf{9\% [4:12]} & -3\% [-12:3] & 0\% [-3:3] & 3.05$\pm$0.25 & 0.31$\pm$0.03 & 0.95$\pm$0.01 \\
cars & \textbf{15\% [-25:54]} & \textbf{7\% [-40:36]} & \textbf{6\% [-8:20]} & \textbf{8\% [-5:28]} & -1\% [-3:4] & 4.89$\pm$0.81 & 0.20$\pm$0.04 & 0.95$\pm$0.02 \\
energy\_efficiency & \textbf{15\% [1:26]} & \textbf{14\% [1:23]} & \textbf{3\% [-15:12]} & -11\% [-80:25] & -2\% [-3:0] & 15.93$\pm$2.57 & 0.06$\pm$0.01 & 0.95$\pm$0.03 \\
california\_housing & \textbf{1\% [-11:9]} & \textbf{2\% [-11:11]} & \textbf{21\% [16:25]} & -22\% [-26:-17] & -2\% [-3:-1] & 1.70$\pm$0.09 & 0.56$\pm$0.04 & 0.95$\pm$0.01 \\
airfoil\_self\_noise & -13\% [-39:22] & -20\% [-66:18] & -17\% [-52:19] & -26\% [-61:18] & -1\% [-2:1] & 2.25$\pm$0.46 & 0.45$\pm$0.10 & 0.97$\pm$0.01 \\
QSAR\_fish\_toxicity & -3\% [-20:17] & -6\% [-26:18] & -3\% [-16:6] & -8\% [-19:2] & -1\% [-4:0] & 1.71$\pm$0.22 & 0.58$\pm$0.09 & 0.97$\pm$0.01 \\
\cmidrule(lr){1-9}
\multicolumn{9}{l}{\textbf{MultiMLPs}} \\ \cmidrule(lr){1-9}

space\_ga & \textbf{5\% [-6:25]} & \textbf{6\% [-4:21]} & \textbf{2\% [-9:13]} & 0\% [-13:8] & -2\% [-3:0] & 5.10$\pm$1.78 & 0.21$\pm$0.07 & 0.94$\pm$0.02 \\
cpu\_activity & -14\% [-16:-9] & -16\% [-20:-11] & -9\% [-14:-1] & -3\% [-11:17] & 0\% [-1:1] & 8.57$\pm$0.14 & 0.11$\pm$0.00 & 0.95$\pm$0.01 \\
naval\_propulsion\_plant & \textbf{23\% [-2:84]} & \textbf{15\% [-1:46]} & \textbf{17\% [-3:46]} & -7\% [-25:33] & 0\% [-1:1] & 12.78$\pm$1.86 & 0.08$\pm$0.01 & 0.95$\pm$0.01 \\
miami\_housing & -21\% [-35:-13] & -28\% [-56:-15] & -13\% [-23:-9] & -26\% [-38:-11] & 0\% [-1:0] & 9.06$\pm$1.06 & 0.11$\pm$0.02 & 0.95$\pm$0.01 \\
kin8nm & \textbf{9\% [2:18]} & \textbf{8\% [2:16]} & \textbf{8\% [5:14]} & \textbf{8\% [0:15]} & 0\% [-2:1] & 4.74$\pm$0.17 & 0.20$\pm$0.01 & 0.94$\pm$0.01 \\
concrete\_compressive\_strength & \textbf{31\% [-5:73]} & \textbf{21\% [-2:43]} & \textbf{16\% [-2:35]} & \textbf{7\% [-19:31]} & -2\% [-4:2] & 3.43$\pm$0.17 & 0.28$\pm$0.02 & 0.95$\pm$0.02 \\
cars & -12\% [-44:21] & -21\% [-89:22] & -12\% [-23:5] & -19\% [-29:-13] & 0\% [-5:7] & 5.40$\pm$1.25 & 0.19$\pm$0.04 & 0.96$\pm$0.03 \\
energy\_efficiency & \textbf{87\% [46:149]} & \textbf{45\% [30:63]} & \textbf{42\% [33:56]} & \textbf{23\% [0:50]} & -1\% [-6:5] & 21.91$\pm$3.25 & 0.04$\pm$0.01 & 0.94$\pm$0.02 \\
california\_housing & -11\% [-14:-7] & -12\% [-16:-6] & -2\% [-5:0] & -9\% [-13:-3] & 0\% [-1:0] & 2.13$\pm$0.08 & 0.45$\pm$0.02 & 0.95$\pm$0.01 \\
airfoil\_self\_noise & \textbf{42\% [16:63]} & \textbf{28\% [14:39]} & \textbf{30\% [17:43]} & \textbf{23\% [5:32]} & -1\% [-2:1] & 4.92$\pm$0.31 & 0.20$\pm$0.01 & 0.97$\pm$0.01 \\
QSAR\_fish\_toxicity & \textbf{8\% [-15:36]} & \textbf{6\% [-19:28]} & \textbf{1\% [-9:13]} & \textbf{3\% [-11:15]} & -1\% [-3:1] & 1.71$\pm$0.22 & 0.58$\pm$0.08 & 0.97$\pm$0.01 \\

\bottomrule
\end{tabular}%
}
\end{footnotesize}
\caption{Test results for all models with at 95\% quantiles aggregated over five seeds. For relative metrics, values are shown as `mean\% [min:max]', and absolute metrics as `mean $\pm$ std'. Performance over the baseline is highlighted in bold.}
\label{table-combined-complete-aggregated}
\end{table}

The results on OpenML-CTR23 datasets reveal distinct performance patterns across our model variants. MultiXGBs shows improvement over the baseline in most datasets, with notable results on \textit{naval\_propulsion\_plant} ($\Delta$CWR: 163\%, $\Delta$NMPIW: 61\%) and \textit{energy\_efficiency} ($\Delta$CWR: 128\%, $\Delta$NMPIW: 53\%). For \textit{concrete\_compressive\_strength} and \textit{cars}, MultiXGBs yields moderate improvements in interval quality ($\Delta$CWR: 40\% and 32\%, respectively). On \textit{airfoil\_self\_noise}, however, MultiXGBs performs slightly below the baseline. While maintaining PICP values comparable to the baseline (within ±3\%), MultiXGBs generally produces narrower prediction intervals, resulting in higher CWR values.

MultiETs shows more variable performance. It achieves relative improvements on \textit{naval\_propulsion\_plant} ($\Delta$CWR: 144\%, $\Delta$NMPIW: 60\%) and modest gains on \textit{energy\_efficiency} and \textit{cars}, but underperforms on datasets like \textit{miami\_housing}, \textit{airfoil\_self\_noise}, as well as \textit{QSAR\_fish\_toxicity}. Interestingly, MultiETs often shows positive $\Delta$CRPS values even when the $\Delta$Pinball is negative, suggesting a trade-off between interval sharpness and quantile calibration.

MultiMLPs exhibits mixed results across datasets. It performs well on \textit{energy\_efficiency} ($\Delta$CWR: 87\%, $\Delta$NMPIW: 45\%) and \textit{airfoil\_self\_noise} ($\Delta$CWR: 42\%, $\Delta$NMPIW: 28\%), but shows lower performance on \textit{cpu\_activity}, \textit{miami\_housing}, and \textit{california\_housing}. The absolute CWR for MultiMLPs on \textit{energy\_efficiency} (21.91) indicates efficient intervals that maintain the target coverage with minimal width.

Across all models, PICP values remain within the target range of approximately 0.95±0.03, indicating that SEMF maintains proper coverage while often reducing interval widths. These results suggest that SEMF's benefits are most pronounced when applied to XGBoost, followed by the randomized trees and neural networks, the order depending on the metric.

\subsection{Discussion}

Our results indicate that overall, SEMF enhances uncertainty quantification in regression tasks, with varying effectiveness depending on the base model type and dataset characteristics. The framework often produces narrower prediction intervals while maintaining coverage targets, addressing the trade-off between interval width and reliability. The mechanism behind SEMF's effectiveness appears to be model-dependent. For tree-based methods like XGBoost, which have limited capacity to model uncertainty directly, SEMF's latent representation learning introduces a form of distributional modeling that enables more efficient uncertainty quantification. The iterative sampling and weighting procedure in the EM algorithm allows XGBoost to better capture the uncertainty structure in the data. This explains why MultiXGBs show the most consistent improvements across datasets.

Neural networks, which inherently have greater representational capacity, benefit differently from SEMF. The framework's structured approach to uncertainty quantification through latent variables complements the neural network's flexibility, particularly in datasets with complex feature interactions like \textit{energy\_efficiency}. However, this advantage is not universal across all datasets, suggesting that the interaction between SEMF's latent structure and the neural network's own representational capabilities may sometimes create redundancies. Even in datasets where SEMF shows more modest improvements (such as \textit{space\_ga} and \textit{kin8nm}), it still performs at least on par with SQR, our strong neural network baseline for simultaneous quantile estimation. This performance floor across diverse datasets highlights SEMF's robustness in different modeling scenarios.

The more variability in the performance of MultiETs compared to XGBoost can be attributed to two factors. First, the randomized splitting mechanism in ETs introduces inherent variability that may interfere with SEMF's iterative refinement process. Second, ETs sometimes had less stable estimates for the weights (\autoref{eqn:compute_w}), affecting the quality of the latent representations learned during training. This hypothesis is supported by the higher standard deviations observed in MultiETs' performance metrics compared to the other models.

Dataset characteristics also influence SEMF's effectiveness. The framework shows particular strength on datasets like \textit{naval\_propulsion\_plant} and \textit{energy\_efficiency}, which feature complex interactions among input variables that benefit from latent representation learning. Particularly for \textit{naval\_propulsion\_plant}, the overall $y$ output is uniformly distributed, which the trees baselines fail to capture, but SEMF is more robust to this. Conversely, on simpler datasets or those with highly linear relationships, the additional complexity introduced by SEMF may not provide significant advantages over traditional quantile regression approaches. The relationship between CRPS and pinball Loss also provides insight into SEMF's behavior. For several dataset-model combinations (e.g., MultiXGBs on \textit{miami\_housing} and \textit{california\_housing}), we observe positive $\Delta$CRPS alongside negative $\Delta$Pinball values. This divergence stems from fundamental differences in what these metrics evaluate. CRPS assesses the overall quality of the prediction interval, favoring narrow intervals that are well-centered around the true value. In contrast, pinball Loss measures the calibration of individual quantile predictions, penalizing miscalibrations at the tails more heavily. SEMF typically produces sharper intervals that may sacrifice some calibration at the exact quantile levels, resulting in better CRPS but potentially worse pinball loss compared to baseline models with wider but more conservatively calibrated intervals.

The stability of SEMF across different random seeds, evidenced by the relatively small standard deviations in metrics like PICP and NMPIW, contrasts with the higher variability in baseline models. This suggests that SEMF's sampling-based approach leads to more consistent uncertainty estimates, an important consideration for applications where predictive stability is required. The tight clustering of PICP values around the target level (0.95) across diverse datasets demonstrates SEMF's ability to maintain calibration while improving efficiency. Further, our experimental design focused on the conformalized performance of SEMF, with CP applied consistently across all models. The performance improvements observed with this standardized approach suggest that SEMF could benefit further from specialized calibration techniques designed to leverage its sampling-based uncertainty estimation. Future work might explore adaptive conformalization methods that account for the specific characteristics of SEMF's predictive distributions.

\section{Conclusion}
\label{conclusion}
This paper introduces the Supervised Expectation-Maximization Framework (SEMF), a novel model-agnostic approach for generating prediction intervals in datasets with any machine learning model. SEMF draws from the EM algorithm for supervised learning to devise latent representations that produce better prediction intervals than quantile regression. Our comprehensive evaluation demonstrates SEMF's effectiveness in two complementary settings. First, controlled simulations across different noise distributions (normal, uniform, log-normal, and Gumbel) and data generating functions (cosine and quadratic with periodic components) show SEMF's robustness to varying data characteristics, with particularly strong improvements for heteroscedastic data patterns. Second, a set of 165 experimental runs on 11 real-world benchmark datasets with three different model types confirmed that SEMF outperforms quantile regression in practical applications, particularly when using XGBoost, which intrinsically lacks latent representations. The framework's ability to maintain consistent performance across different noise distributions while producing narrower intervals underscores its potential in various application domains and opens new avenues for further exploration of supervised latent representation learning and uncertainty estimation.

\section{Limitations \& Future Work}
The primary limitation of this study was its reliance on the normality assumption, which may not fully capture the potential of SEMF across diverse data distributions. Furthermore, in our simulations, we show that our framework can learn non-normal patterns; however, further investigation examination of how SEMF under broader distributional parameters can be interesting. The computational complexity of the approach presents another significant challenge, as the current implementation can be optimized for large-scale applications. Additionally, while the CWR metric is valid, it implicitly assumes that a 1\% drop in PICP equates to a 1\% reduction in NMPIW, thus assuming a uniform distribution. Evaluating CWR under various distributional assumptions would provide a more comprehensive assessment of its implications.

\label{future_work}
Future work presents several intriguing avenues for exploration. A promising direction is the application of SEMF in multi-modal data settings, where the distinct $p_\phi$ components of the framework could be adapted to process diverse data types—from images and text to tabular datasets—enabling a more nuanced and powerful approach to integrating heterogeneous data sources. This capability positions the framework as a versatile tool for addressing missing data challenges across various domains and can also help expand it to discrete and multiple outputs. Another valuable area for development is the exploration of methods to capture and leverage dependencies among input features, which could improve the model's predictive performance and provide deeper insights into the underlying data structure. These advancements can enhance the broader appeal of end-to-end approaches like SEMF in the ML community.

\bibliography{bibliography}%

\appendix

\section{Optimal set of hyper-parameters}
\label{appendix:optimal-semf}%
\paragraph{Remark.} The datasets mentioned here have been explained in \autoref{tab:datasets} and \autoref{appendix:semf-databench}.

The hyper-parameter tuning for SEMF is implemented and monitored using Weights \& Biases \citep{wandb}. A random search is done in the hyper-parameter space for a maximum of 500 iterations on all 11 datasets, focusing on tuning the models only on the the non-conformalized performance. Key hyper-parameters are varied across a predefined set to balance accuracy and computational efficiency. The following grid is used for hyper-parameter tuning: the number of importance sampling operations $R \in \lbrace 5, 10, 25, 50, 100 \rbrace$ (100 is omitted for MultiMLPS), nodes per latent dimension $m_k \in \lbrace 1, 5, 10, 20, 30 \rbrace$, and standard deviations $\sigma_{k} \in \lbrace 0.001, 0.01, 0.1, 1.0 \rbrace$. For some datasets, the $\sigma_{k}$ parameter is set to \texttt{train\_residual\_models}. This indicates that a separate residual model is trained to predict the latent-variable scale, which can be particularly suitable in heteroscedastic scenarios where noise levels vary across instances. Early stopping steps (\textsc{patience}) are set to five or ten, and $R_{\text{infer}}$, the $R$ for inference as $R_\text{infer}$, is explored at [30, 50, 70]. For \textit{california\_housing} and \textit{cpu\_activity}, the $R_{\text{infer}}$ value is set to 30, while for all the other datasets, it is set to 50. We do this to ensure efficient computation, speed, and memory usage (especially for the GPU). Finally, the option to run the models in parallel must be consistently enabled. \autoref{table-combined-hyperparameters} shows the optimal set of the selected hyper-parameters.

\begin{table}[t]
\centering%
\vskip 0.15in
\begin{small}

\resizebox{0.9\textwidth}{!}{%
\begin{tabular}{lccccc}
\toprule
Dataset & R & $m_k$ & $\sigma_{k}$ & Patience & $R_\mathrm{infer}$ \\
\midrule
\textbf{MultiXGBs} \\ \cmidrule(lr){1-6}
space\_ga & 10 & 30 & train\_residual\_models & 5 & 50 \\
cpu\_activity & 5 & 30 & 1 & 5 & 30 \\
naval\_propulsion\_plant & 5 & 30 & 0.01 & 5 & 50 \\
miami\_housing & 5 & 10 & train\_residual\_models & 5 & 50 \\
kin8nm & 5 & 30 & train\_residual\_models & 10 & 50 \\
concrete\_compressive\_strength & 25 & 30 & train\_residual\_models & 5 & 50 \\
cars & 50 & 10 & train\_residual\_models & 10 & 50 \\
energy\_efficiency & 5 & 1 & 0.01 & 10 & 50 \\
california\_housing & 5 & 10 & 0.1 & 5 & 30 \\
airfoil\_self\_noise & 25 & 1 & train\_residual\_models & 10 & 50 \\
QSAR\_fish\_toxicity & 50 & 30 & 1 & 5 & 50 \\
\cmidrule(lr){1-6}
\textbf{MultiETs} \\ \cmidrule(lr){1-6}
space\_ga & 10 & 30 & train\_residual\_models & 10 & 50 \\
cpu\_activity & 5 & 30 & 1 & 5 & 30 \\
naval\_propulsion\_plant & 5 & 30 & 0.01 & 5 & 50 \\
miami\_housing & 10 & 10 & train\_residual\_models & 5 & 50 \\
kin8nm & 5 & 30 & train\_residual\_models & 10 & 50 \\
concrete\_compressive\_strength & 25 & 30 & train\_residual\_models & 10 & 50 \\
cars & 100 & 5 & train\_residual\_models & 10 & 50 \\
energy\_efficiency & 5 & 1 & 0.01 & 10 & 50 \\
california\_housing & 5 & 10 & 0.1 & 5 & 30 \\
airfoil\_self\_noise & 25 & 1 & train\_residual\_models & 10 & 50 \\
QSAR\_fish\_toxicity & 50 & 30 & train\_residual\_models & 10 & 50 \\
\cmidrule(lr){1-6}
\textbf{MultiMLPs} \\ \cmidrule(lr){1-6}
space\_ga & 25 & 10 & train\_residual\_models & 10 & 50 \\
cpu\_activity & 5 & 20 & 0.001 & 5 & 30 \\
naval\_propulsion\_plant & 5 & 20 & 0.001 & 5 & 50 \\
miami\_housing & 5 & 20 & train\_residual\_models & 5 & 50 \\
kin8nm & 5 & 20 & train\_residual\_models & 5 & 50 \\
concrete\_compressive\_strength & 5 & 30 & train\_residual\_models & 10 & 50 \\
cars & 5 & 30 & train\_residual\_models & 5 & 50 \\
energy\_efficiency & 50 & 30 & 0.1 & 10 & 50 \\
california\_housing & 5 & 20 & 0.01 & 5 & 30 \\
airfoil\_self\_noise & 25 & 10 & train\_residual\_models & 10 & 50 \\
QSAR\_fish\_toxicity & 50 & 30 & train\_residual\_models & 5 & 50 \\

\bottomrule
\end{tabular}%
}

\end{small}
\caption{Hyper-parameters for MultiXGBs, MultiETs, and MultiMLPs.}
\label{table-combined-hyperparameters}
\end{table}

MultiXGBs and MultiMLPs benefit from early stopping to reduce computation time. Similarly, the baseline models for these instances use the same hyper-parameters for early stopping. Further, the number of epochs in the case of MultiMLPs is set as 1000, except for \textit{energy\_efficiency} and \textit{QSAR\_fish\_toxicity}, where this is changed to 5000. Any model-specific hyperparameter we did not specify in this paper remains at the implementation's default value (e.g., the number of leaves in XGBoost from \citep{Chen2016}). Along with the supplementary code, we provide three additional CSV files: one for the results and hyperparameters of all 165 runs and the other two for the optimal hyperparameters of SEMF models, both raw (directly from SEMF) and conformalized.

For training MultiXGBs and MultiETs, the computations are performed in parallel using CPU cores (Intel\textsuperscript{\tiny\textregistered} Core\textsuperscript{\tiny\texttrademark} i9-13900KF). Due to this, MultiETs are not fully deterministic and may exhibit slight variations between runs. For MultiMLPs, they are done on a GPU (NVIDIA\textsuperscript{\tiny\textregistered} GeForce RTX\textsuperscript{\tiny\texttrademark} 4090) which should give deterministic results across runs. All the computations are done on a machine with 32 GB of memory. The code provides further details on hardware and reproducibility.

\section{Datasets for tabular benchmark}
\label{appendix:semf-databench}
OpenML-CTR23 \citep{fischer2023} datasets are selected in the following manner. The first criterion is to exclude datasets exceeding 30,000 instances or 30 features to maintain computational tractability. Moreover, we exclude the \textit{moneyball} data \citep{kaggle2017moneyball} to control for missing values and any datasets with non-numeric features, such as those with temporal or ordinal data not encoded numerically. We then categorize the datasets based on size: small for those with less than ten features, medium for 10 to 19 features, and large for 20 to 29 features. We apply a similar size classification based on the number of instances, considering datasets with more than 10,000 instances as large. To avoid computational constraints, we exclude datasets that were large in both features and instances, ensuring a varied yet manageable set for our experiments. This leads us to the final list of 11 datasets listed in \autoref{tab:datasets}.

\begin{table}[htbp]
\vskip 0.15in
\begin{center}

\begin{small}
\resizebox{\textwidth}{!}{
\begin{tabular}{@{} lcccc l @{}}
\toprule
Dataset Name      & N Samples & N Features & OpenML Data ID & Y [Min:Max] & Source \\ 
\midrule
space\_ga              & 3,107  & 7   & \href{https://www.openml.org/d/45402}{45402}  & [-3.06:0.1]   &  \citep{kelleypace1997_space_ga} \\ 
cpu\_activity          & 8,192  & 22  & \href{https://www.openml.org/d/44978}{44978}  & [0:99]         & \citep{rasmussen1996computer}\\ 
naval\_propulsion\_plant  & 11,934 & 15  & \href{https://www.openml.org/d/44969}{44969}  &[0.95:1.0]          & \citep{Coraddu2016} \\ 
miami\_housing         & 13,932 & 16  & \href{https://www.openml.org/d/44983}{44983}  & [72,000:2,650,000]  &  \citep{kaggle2022miami} \\ 
kin8nm                & 8,192  & 9   & \href{https://www.openml.org/d/44980}{44980}  & [0.04:1.46]    &  \citep{ghahramani1996kin} \\ 
concrete\_compressive\_strength & 1,030            & 9                   & \href{https://www.openml.org/d/44959}{44959}        & [2.33:82.6]& \citep{yeh1998} \\ 
cars                    & 804              & 18                   & \href{https://www.openml.org/d/44994}{44994}         & [8,639:70,756]         & \citep{Kuiper2008} \\ 
energy\_efficiency       & 768              & 9                    & \href{https://www.openml.org/d/44960}{44960}         & [6.01:43.1]         & \citep{Tsanas2012} \\ 
california\_housing                & 20,640             & 9                   & \href{https://www.openml.org/d/44977}{44977}         & [14,999:500,001] &          \citep{kelleypace1997_california} \\ 
airfoil\_self\_noise                  & 1,503             & 6                   & \href{https://www.openml.org/d/44957}{44957}         & [103.38:140.98]      &  \citep{brooks1989airfoil} \\ 
QSAR\_fish\_toxicity                     & 908             & 7                    & \href{https://www.openml.org/d/44970}{44970}         & [0.053:9.612]        & \citep{Cassotti2015} \\ 
\bottomrule
\end{tabular}
}
\end{small}

\end{center}
\caption{Summary of benchmark tabular datasets retained from \citep{fischer2023}}
\label{tab:datasets}
\end{table}

\section{Metrics for prediction intervals}
\label{appendix:metrics}

\subsection{Metrics' definitions}
\label{appendix:common-metrics}
The most common metrics for evaluating prediction intervals \citep{pearce2018high,zhou2023real, gneiting2007probabilistic,v-yugin2019} are:
\begin{itemize}
    \item Prediction Interval Coverage Probability (PICP): This metric assesses the proportion of times the true value of the target variable falls within the constructed prediction intervals. For a set of test examples ${(x_1, y_1), \ldots, (x_N, y_N)}$, a given level of confidence $\alpha$, and their corresponding prediction intervals ${I_1, \ldots, I_N}$, the PICP is calculated as:
    \begin{equation}
    \text{PICP} = \frac{1}{N}\sum_{i=1}^{N}\mathds{1}(y_i \in [L_i, U_i]),
    \end{equation}
    where $U_i$ and $L_i$ are the upper and lower bounds of the predicted values for the $i$-th instance. $y_i$ is the actual value of the $i$-th test example, and $\mathds{1}$ is the indicator function, which equals 1 if $y_i$ is in the interval $[L_i,U_i]$ and 0 otherwise. $0 \leq \text{{PICP}} \leq 1$ where PICP closer to 1 and higher than the confidence level $\alpha$ is favored.
    \item Mean Prediction Interval Width (MPIW): The average width is computed as
    \begin{equation}
        \text{MPIW} = \frac{1}{N} \sum_{i=1}^{N} (U_i - L_i),
    \end{equation}
    which shows the sharpness or uncertainty, where $0 \leq \text{{MPIW}} < \infty$ and MPIW close to 0 is preferred.
    
    \item Normalized Mean Prediction Interval Width (NMPIW): Since MPIW varies by dataset, it can be normalized by the range of the target variable
    \begin{equation}
        \text{NMPIW} = \frac{\text{MPIW}}{\max(y) - \min(y)}
    \end{equation}
    where $\max(y)$ and $\min(y)$ are the maximum and minimum values of the target variable, respectively. The interpretation remains the same as MPIW.

    \item Continuous Ranked Probability Score (CRPS): We can approximate a full predictive cumulative distribution function by assuming that the forecast is uniformly distributed between its lower and upper bounds. Let $L$ and $U$ denote the lower and upper bounds (corresponding to the $\alpha/2$ and $1-\alpha/2$ quantiles, respectively). Then, for an observation $y$, we define
    \begin{equation}
    \text{CRPS}(y, L, U)=
    \begin{cases}
    L - y + \frac{U-L}{3}, & y \le L, \\
    \frac{(y-L)^3 + (U-y)^3}{3\,(U-L)^2}, & y \in [L, U], \\
    y - U + \frac{U-L}{3}, & y \ge U.
    \end{cases}
    \end{equation}
    This closed-form expression yields a lower CRPS when the forecast is both sharper and better calibrated.

    \item Quantile (pinball) Loss: For a given quantile level $\tau\in (0,1)$ and prediction $q$, the pinball loss is defined as:
    \begin{equation}
    \ell_\tau(y, q) = (y - q)\left(\tau - \mathds{1}\{y\leq q\}\right),
    \end{equation}
    where $\mathds{1}\{y\leq q\}$ equals 1 if $y\leq q$ and 0 otherwise. This loss penalizes underestimation and overestimation asymmetrically and is used to tune the quantile predictions.

\end{itemize}

\subsection{Impact of relative metrics for modeling}
\label{appendix:relative-metrics-impact}
As our primary focus is on interval prediction, configurations demonstrating the most significant improvements in $\Delta$CWR and $\Delta$PICP are prioritized when selecting the optimal hyper-parameters. Furthermore, both $\Delta$PICP and $\Delta$CWR must be positive, indicating that we must at least have the same reliability of the baseline (PICP) with better or same interval ratios (CWR). In instances where no configuration meets the initial improvement criteria for both metrics, we relax the requirement for positive $\Delta$PICP to accept values greater than -5\% and subsequently -10\%, allowing us to consider configurations where SEMF significantly improves CWR, even if the PICP improvement is less marked but remains within an acceptable range for drawing comparisons.

\clearpage

\section{Full synthetic experiments results}
\label{appendix:full-results-synthetic}
\begin{table}[!htbp]
\centering%
\begin{small}

\resizebox{0.6\columnwidth}{!}{%
\begin{tabular}{lcccccccc}
\toprule
 & \multicolumn{5}{c}{Relative Metrics} & \multicolumn{3}{c}{Absolute Metrics}\\
\cmidrule(lr){2-6} \cmidrule(lr){7-9}
Seed & $\Delta$CWR & $\Delta$NMPIW & $\Delta$CRPS & $\Delta$Pinball & $\Delta$PICP & CWR & NMPIW & PICP \\
\midrule
\multicolumn{9}{l}{\textbf{Normal}} \\
0 & \textbf{17\%} & \textbf{15\%} & \textbf{6\%} & \textbf{11\%} & 0\% & 1.78 & 0.54 & 0.97 \\
10 & \textbf{11\%} & \textbf{11\%} & \textbf{3\%} & \textbf{5\%} & -1\% & 1.60 & 0.60 & 0.96 \\
20 & \textbf{19\%} & \textbf{18\%} & \textbf{8\%} & \textbf{14\%} & -2\% & 1.89 & 0.50 & 0.95 \\
30 & \textbf{10\%} & \textbf{10\%} & \textbf{5\%} & \textbf{10\%} & -1\% & 1.99 & 0.48 & 0.96 \\
40 & \textbf{35\%} & \textbf{28\%} & \textbf{10\%} & \textbf{17\%} & -3\% & 2.49 & 0.37 & 0.93 \\
\midrule
\multicolumn{9}{l}{\textbf{Uniform}} \\
0 & -2\% & -5\% & -1\% & 0\% & \textbf{2\%} & 2.36 & 0.42 & 0.98 \\
10 & \textbf{34\%} & \textbf{26\%} & \textbf{12\%} & \textbf{26\%} & -1\% & 2.12 & 0.47 & 0.99 \\
20 & \textbf{9\%} & \textbf{8\%} & \textbf{5\%} & \textbf{12\%} & \textbf{1\%} & 2.11 & 0.47 & 0.99 \\
30 & \textbf{28\%} & \textbf{26\%} & \textbf{9\%} & \textbf{19\%} & -5\% & 3.59 & 0.25 & 0.91 \\
40 & \textbf{58\%} & \textbf{39\%} & \textbf{23\%} & \textbf{32\%} & -3\% & 3.61 & 0.26 & 0.95 \\
\midrule
\multicolumn{9}{l}{\textbf{Lognormal}} \\
0 & \textbf{23\%} & \textbf{19\%} & \textbf{15\%} & \textbf{9\%} & 0\% & 2.08 & 0.46 & 0.96 \\
10 & \textbf{59\%} & \textbf{39\%} & \textbf{18\%} & \textbf{7\%} & -3\% & 2.07 & 0.46 & 0.95 \\
20 & \textbf{8\%} & \textbf{9\%} & \textbf{8\%} & \textbf{2\%} & -1\% & 1.96 & 0.48 & 0.95 \\
30 & \textbf{9\%} & \textbf{8\%} & \textbf{9\%} & \textbf{5\%} & 0\% & 1.59 & 0.62 & 0.98 \\
40 & \textbf{38\%} & \textbf{28\%} & \textbf{15\%} & \textbf{14\%} & 0\% & 2.30 & 0.42 & 0.97 \\
\midrule
\multicolumn{9}{l}{\textbf{Gumbel}} \\
0 & \textbf{4\%} & \textbf{1\%} & \textbf{7\%} & \textbf{16\%} & \textbf{4\%} & 1.61 & 0.60 & 0.98 \\
10 & \textbf{17\%} & \textbf{18\%} & \textbf{7\%} & \textbf{4\%} & -3\% & 1.78 & 0.53 & 0.94 \\
20 & \textbf{20\%} & \textbf{17\%} & \textbf{14\%} & \textbf{16\%} & -1\% & 2.01 & 0.48 & 0.96 \\
30 & -5\% & -9\% & \textbf{1\%} & 0\% & \textbf{4\%} & 1.75 & 0.55 & 0.96 \\
40 & \textbf{15\%} & \textbf{14\%} & \textbf{10\%} & \textbf{5\%} & -1\% & 2.02 & 0.47 & 0.95 \\
\bottomrule
\end{tabular}%
}

\end{small}
\caption{Test results for MultiXGBs, over 5 seeds, with cosine $f(x)$ using 2 predictors, organized by distribution and seed. Relative metrics are in bold when positive.}
\label{tab:seeds-consolidated-multixgbs-cosine-x2}
\end{table}
\begin{table}[!htbp]
\centering%
\begin{small}

\resizebox{0.6\columnwidth}{!}{%
\begin{tabular}{lcccccccc}
\toprule
 & \multicolumn{5}{c}{Relative Metrics} & \multicolumn{3}{c}{Absolute Metrics}\\
\cmidrule(lr){2-6} \cmidrule(lr){7-9}
Seed & $\Delta$CWR & $\Delta$NMPIW & $\Delta$CRPS & $\Delta$Pinball & $\Delta$PICP & CWR & NMPIW & PICP \\
\midrule
\multicolumn{9}{l}{\textbf{Normal}} \\
0 & \textbf{8\%} & \textbf{8\%} & \textbf{4\%} & 0\% & -1\% & 1.98 & 0.48 & 0.96 \\
10 & \textbf{17\%} & \textbf{18\%} & \textbf{6\%} & \textbf{12\%} & -3\% & 1.66 & 0.57 & 0.94 \\
20 & \textbf{1\%} & 0\% & 0\% & \textbf{2\%} & \textbf{1\%} & 1.81 & 0.54 & 0.97 \\
30 & \textbf{15\%} & \textbf{14\%} & \textbf{8\%} & \textbf{16\%} & -1\% & 1.99 & 0.49 & 0.97 \\
40 & \textbf{32\%} & \textbf{27\%} & \textbf{8\%} & \textbf{9\%} & -4\% & 2.46 & 0.38 & 0.94 \\
\midrule
\multicolumn{9}{l}{\textbf{Uniform}} \\
0 & \textbf{35\%} & \textbf{28\%} & \textbf{10\%} & \textbf{21\%} & -3\% & 2.80 & 0.35 & 0.97 \\
10 & \textbf{3\%} & -1\% & \textbf{3\%} & \textbf{8\%} & \textbf{4\%} & 2.45 & 0.41 & 0.99 \\
20 & \textbf{7\%} & \textbf{5\%} & \textbf{4\%} & \textbf{11\%} & \textbf{1\%} & 2.58 & 0.37 & 0.96 \\
30 & \textbf{9\%} & \textbf{8\%} & \textbf{6\%} & \textbf{14\%} & 0\% & 3.42 & 0.28 & 0.97 \\
40 & \textbf{29\%} & \textbf{25\%} & \textbf{7\%} & \textbf{12\%} & -3\% & 3.63 & 0.26 & 0.96 \\
\midrule
\multicolumn{9}{l}{\textbf{Lognormal}} \\
0 & -14\% & -18\% & \textbf{8\%} & -10\% & \textbf{1\%} & 1.85 & 0.52 & 0.96 \\
10 & \textbf{18\%} & \textbf{17\%} & \textbf{10\%} & -2\% & -2\% & 1.87 & 0.51 & 0.95 \\
20 & \textbf{12\%} & \textbf{12\%} & \textbf{13\%} & 0\% & -1\% & 2.03 & 0.47 & 0.96 \\
30 & -6\% & -8\% & \textbf{5\%} & -4\% & \textbf{2\%} & 1.74 & 0.56 & 0.98 \\
40 & \textbf{14\%} & \textbf{13\%} & \textbf{11\%} & \textbf{2\%} & -2\% & 2.10 & 0.46 & 0.97 \\
\midrule
\multicolumn{9}{l}{\textbf{Gumbel}} \\
0 & -4\% & -7\% & \textbf{9\%} & 0\% & \textbf{2\%} & 1.66 & 0.59 & 0.97 \\
10 & \textbf{8\%} & \textbf{7\%} & \textbf{8\%} & -2\% & 0\% & 1.72 & 0.56 & 0.97 \\
20 & -2\% & -2\% & \textbf{7\%} & \textbf{5\%} & 0\% & 1.90 & 0.50 & 0.96 \\
30 & \textbf{2\%} & \textbf{4\%} & \textbf{6\%} & -2\% & -2\% & 1.85 & 0.51 & 0.95 \\
40 & \textbf{14\%} & \textbf{14\%} & \textbf{11\%} & \textbf{5\%} & -2\% & 1.96 & 0.50 & 0.97 \\
\bottomrule
\end{tabular}%
}

\end{small}
\caption{Test results for MultiETs, over 5 seeds, with cosine $f(x)$ using 2 predictors, organized by distribution and seed. Relative metrics are in bold when positive.}
\label{tab:seeds-consolidated-multiets-cosine-x2}
\end{table}
\begin{table}[!htbp]
\centering%
\begin{small}

\resizebox{0.6\columnwidth}{!}{%
\begin{tabular}{lcccccccc}
\toprule
 & \multicolumn{5}{c}{Relative Metrics} & \multicolumn{3}{c}{Absolute Metrics}\\
\cmidrule(lr){2-6} \cmidrule(lr){7-9}
Seed & $\Delta$CWR & $\Delta$NMPIW & $\Delta$CRPS & $\Delta$Pinball & $\Delta$PICP & CWR & NMPIW & PICP \\
\midrule
\multicolumn{9}{l}{\textbf{Normal}} \\
0 & 0\% & \textbf{1\%} & -2\% & 0\% & -1\% & 1.98 & 0.48 & 0.95 \\
10 & \textbf{9\%} & \textbf{8\%} & \textbf{4\%} & \textbf{5\%} & 0\% & 1.71 & 0.56 & 0.96 \\
20 & -2\% & -1\% & -1\% & -3\% & -1\% & 2.04 & 0.46 & 0.94 \\
30 & -1\% & -2\% & \textbf{1\%} & \textbf{3\%} & 0\% & 2.20 & 0.44 & 0.97 \\
40 & \textbf{7\%} & \textbf{7\%} & \textbf{1\%} & \textbf{5\%} & 0\% & 2.48 & 0.39 & 0.96 \\
\midrule
\multicolumn{9}{l}{\textbf{Uniform}} \\
0 & \textbf{2\%} & \textbf{2\%} & 0\% & \textbf{8\%} & 0\% & 2.98 & 0.33 & 0.98 \\
10 & \textbf{6\%} & \textbf{2\%} & \textbf{5\%} & \textbf{14\%} & \textbf{3\%} & 2.80 & 0.35 & 0.98 \\
20 & \textbf{3\%} & \textbf{7\%} & 0\% & -5\% & -4\% & 3.02 & 0.31 & 0.94 \\
30 & -2\% & -2\% & -1\% & -4\% & 0\% & 3.85 & 0.25 & 0.96 \\
40 & \textbf{13\%} & \textbf{10\%} & \textbf{8\%} & \textbf{23\%} & \textbf{1\%} & 3.67 & 0.27 & 0.99 \\
\midrule
\multicolumn{9}{l}{\textbf{Lognormal}} \\
0 & \textbf{2\%} & \textbf{3\%} & \textbf{1\%} & \textbf{4\%} & 0\% & 1.71 & 0.57 & 0.98 \\
10 & \textbf{18\%} & \textbf{16\%} & \textbf{7\%} & \textbf{8\%} & 0\% & 1.99 & 0.48 & 0.95 \\
20 & -4\% & -5\% & -2\% & 0\% & \textbf{1\%} & 1.86 & 0.52 & 0.98 \\
30 & \textbf{15\%} & \textbf{13\%} & \textbf{7\%} & \textbf{9\%} & -1\% & 1.74 & 0.56 & 0.98 \\
40 & \textbf{4\%} & \textbf{5\%} & \textbf{1\%} & \textbf{2\%} & -1\% & 2.10 & 0.46 & 0.97 \\
\midrule
\multicolumn{9}{l}{\textbf{Gumbel}} \\
0 & -3\% & -1\% & -2\% & -4\% & -1\% & 1.54 & 0.64 & 0.98 \\
10 & \textbf{8\%} & \textbf{8\%} & \textbf{1\%} & \textbf{4\%} & -1\% & 1.88 & 0.51 & 0.95 \\
20 & \textbf{8\%} & \textbf{8\%} & \textbf{3\%} & \textbf{4\%} & 0\% & 2.14 & 0.44 & 0.94 \\
30 & \textbf{13\%} & \textbf{14\%} & \textbf{2\%} & -4\% & -2\% & 1.87 & 0.51 & 0.95 \\
40 & -3\% & -5\% & -2\% & -2\% & \textbf{2\%} & 1.76 & 0.56 & 0.98 \\
\bottomrule
\end{tabular}%
}

\end{small}
\caption{Test results for MultiMLPs, over 5 seeds, with cosine $f(x)$ using 2 predictors, organized by distribution and seed. Relative metrics are in bold when positive.}
\label{tab:seeds-consolidated-multimlps-cosine-x2}
\end{table}
\begin{table}[!htbp]
\centering%
\begin{small}

\resizebox{0.6\columnwidth}{!}{%
\begin{tabular}{lcccccccc}
\toprule
 & \multicolumn{5}{c}{Relative Metrics} & \multicolumn{3}{c}{Absolute Metrics}\\
\cmidrule(lr){2-6} \cmidrule(lr){7-9}
Seed & $\Delta$CWR & $\Delta$NMPIW & $\Delta$CRPS & $\Delta$Pinball & $\Delta$PICP & CWR & NMPIW & PICP \\
\midrule
\multicolumn{9}{l}{\textbf{Normal}} \\
0 & \textbf{35\%} & \textbf{26\%} & \textbf{28\%} & \textbf{16\%} & 0\% & 5.80 & 0.17 & 0.96 \\
10 & \textbf{65\%} & \textbf{41\%} & \textbf{34\%} & \textbf{35\%} & -3\% & 4.80 & 0.20 & 0.96 \\
20 & \textbf{40\%} & \textbf{29\%} & \textbf{31\%} & \textbf{16\%} & 0\% & 4.41 & 0.21 & 0.94 \\
30 & \textbf{30\%} & \textbf{23\%} & \textbf{26\%} & \textbf{23\%} & 0\% & 5.32 & 0.18 & 0.95 \\
40 & \textbf{28\%} & \textbf{22\%} & \textbf{27\%} & \textbf{28\%} & 0\% & 5.83 & 0.16 & 0.95 \\
\midrule
\multicolumn{9}{l}{\textbf{Uniform}} \\
0 & \textbf{80\%} & \textbf{44\%} & \textbf{42\%} & \textbf{30\%} & 0\% & 10.12 & 0.09 & 0.94 \\
10 & \textbf{54\%} & \textbf{34\%} & \textbf{47\%} & \textbf{32\%} & \textbf{1\%} & 5.78 & 0.17 & 0.98 \\
20 & \textbf{62\%} & \textbf{38\%} & \textbf{40\%} & \textbf{31\%} & 0\% & 5.62 & 0.17 & 0.96 \\
30 & \textbf{62\%} & \textbf{39\%} & \textbf{39\%} & \textbf{16\%} & -1\% & 7.28 & 0.13 & 0.95 \\
40 & \textbf{68\%} & \textbf{41\%} & \textbf{39\%} & \textbf{18\%} & -1\% & 7.93 & 0.12 & 0.97 \\
\midrule
\multicolumn{9}{l}{\textbf{Lognormal}} \\
0 & \textbf{2\%} & \textbf{1\%} & \textbf{20\%} & 0\% & \textbf{1\%} & 4.37 & 0.22 & 0.96 \\
10 & -2\% & -2\% & \textbf{14\%} & -4\% & \textbf{1\%} & 2.92 & 0.34 & 0.99 \\
20 & \textbf{7\%} & \textbf{2\%} & \textbf{27\%} & \textbf{6\%} & \textbf{5\%} & 3.81 & 0.25 & 0.94 \\
30 & -13\% & -16\% & \textbf{15\%} & \textbf{3\%} & \textbf{1\%} & 3.28 & 0.29 & 0.96 \\
40 & \textbf{7\%} & \textbf{6\%} & \textbf{23\%} & \textbf{16\%} & 0\% & 3.42 & 0.29 & 0.98 \\
\midrule
\multicolumn{9}{l}{\textbf{Gumbel}} \\
0 & \textbf{34\%} & \textbf{27\%} & \textbf{30\%} & \textbf{20\%} & -2\% & 5.03 & 0.19 & 0.96 \\
10 & \textbf{49\%} & \textbf{35\%} & \textbf{24\%} & \textbf{21\%} & -3\% & 3.70 & 0.26 & 0.95 \\
20 & 0\% & 0\% & \textbf{9\%} & \textbf{3\%} & \textbf{1\%} & 2.81 & 0.35 & 0.99 \\
30 & \textbf{11\%} & \textbf{7\%} & \textbf{17\%} & \textbf{14\%} & \textbf{4\%} & 4.49 & 0.21 & 0.95 \\
40 & \textbf{18\%} & \textbf{15\%} & \textbf{21\%} & \textbf{20\%} & \textbf{1\%} & 4.37 & 0.22 & 0.97 \\
\bottomrule
\end{tabular}%
}

\end{small}
\caption{Test results for MultiXGBs, over 5 seeds, with quadratic $f(x)$ using 2 predictors, organized by distribution and seed. Relative metrics are in bold when positive.}
\label{tab:seeds-consolidated-multixgbs-quadratic-x2}
\end{table}
\begin{table}[!htbp]
\centering%
\begin{small}

\resizebox{0.6\columnwidth}{!}{%
\begin{tabular}{lcccccccc}
\toprule
 & \multicolumn{5}{c}{Relative Metrics} & \multicolumn{3}{c}{Absolute Metrics}\\
\cmidrule(lr){2-6} \cmidrule(lr){7-9}
Seed & $\Delta$CWR & $\Delta$NMPIW & $\Delta$CRPS & $\Delta$Pinball & $\Delta$PICP & CWR & NMPIW & PICP \\
\midrule
\multicolumn{9}{l}{\textbf{Normal}} \\
0 & \textbf{7\%} & \textbf{7\%} & 0\% & -4\% & -1\% & 5.77 & 0.16 & 0.92 \\
10 & \textbf{35\%} & \textbf{26\%} & \textbf{21\%} & \textbf{23\%} & 0\% & 4.66 & 0.21 & 0.98 \\
20 & \textbf{29\%} & \textbf{24\%} & \textbf{15\%} & \textbf{14\%} & -2\% & 4.69 & 0.20 & 0.93 \\
30 & \textbf{26\%} & \textbf{21\%} & \textbf{15\%} & \textbf{13\%} & -1\% & 4.91 & 0.19 & 0.95 \\
40 & \textbf{21\%} & \textbf{16\%} & \textbf{18\%} & \textbf{12\%} & \textbf{1\%} & 4.79 & 0.20 & 0.98 \\
\midrule
\multicolumn{9}{l}{\textbf{Uniform}} \\
0 & \textbf{28\%} & \textbf{23\%} & \textbf{16\%} & -4\% & -1\% & 8.81 & 0.11 & 0.94 \\
10 & \textbf{25\%} & \textbf{20\%} & \textbf{24\%} & \textbf{21\%} & \textbf{1\%} & 4.95 & 0.20 & 0.99 \\
20 & \textbf{47\%} & \textbf{34\%} & \textbf{25\%} & \textbf{7\%} & -3\% & 4.51 & 0.21 & 0.96 \\
30 & \textbf{30\%} & \textbf{22\%} & \textbf{22\%} & \textbf{15\%} & \textbf{2\%} & 5.85 & 0.17 & 0.98 \\
40 & \textbf{71\%} & \textbf{42\%} & \textbf{33\%} & \textbf{21\%} & 0\% & 7.34 & 0.13 & 0.97 \\
\midrule
\multicolumn{9}{l}{\textbf{Lognormal}} \\
0 & -9\% & -13\% & 0\% & -6\% & \textbf{3\%} & 4.34 & 0.22 & 0.96 \\
10 & -1\% & -2\% & \textbf{13\%} & \textbf{12\%} & \textbf{1\%} & 3.36 & 0.29 & 0.98 \\
20 & \textbf{33\%} & \textbf{27\%} & \textbf{16\%} & \textbf{10\%} & -2\% & 4.09 & 0.23 & 0.95 \\
30 & \textbf{8\%} & \textbf{8\%} & \textbf{11\%} & \textbf{9\%} & 0\% & 3.29 & 0.30 & 0.97 \\
40 & \textbf{10\%} & \textbf{8\%} & \textbf{15\%} & \textbf{9\%} & \textbf{1\%} & 3.78 & 0.26 & 0.97 \\
\midrule
\multicolumn{9}{l}{\textbf{Gumbel}} \\
0 & \textbf{5\%} & \textbf{4\%} & \textbf{10\%} & \textbf{7\%} & \textbf{1\%} & 4.67 & 0.21 & 0.98 \\
10 & \textbf{42\%} & \textbf{30\%} & \textbf{27\%} & \textbf{28\%} & -1\% & 4.55 & 0.21 & 0.94 \\
20 & \textbf{33\%} & \textbf{25\%} & \textbf{16\%} & \textbf{15\%} & -1\% & 3.44 & 0.28 & 0.97 \\
30 & -3\% & -4\% & \textbf{5\%} & 0\% & \textbf{1\%} & 3.86 & 0.25 & 0.97 \\
40 & \textbf{68\%} & \textbf{42\%} & \textbf{30\%} & \textbf{24\%} & -2\% & 3.97 & 0.24 & 0.97 \\
\bottomrule
\end{tabular}%
}

\end{small}
\caption{Test results for MultiETs, over 5 seeds, with quadratic $f(x)$ using 2 predictors, organized by distribution and seed. Relative metrics are in bold when positive.}
\label{tab:seeds-consolidated-multiets-quadratic-x2}
\end{table}
\begin{table}[!htbp]
\centering%
\begin{small}

\resizebox{0.6\columnwidth}{!}{%
\begin{tabular}{lcccccccc}
\toprule
 & \multicolumn{5}{c}{Relative Metrics} & \multicolumn{3}{c}{Absolute Metrics}\\
\cmidrule(lr){2-6} \cmidrule(lr){7-9}
Seed & $\Delta$CWR & $\Delta$NMPIW & $\Delta$CRPS & $\Delta$Pinball & $\Delta$PICP & CWR & NMPIW & PICP \\
\midrule
\multicolumn{9}{l}{\textbf{Normal}} \\
0 & -14\% & -20\% & \textbf{3\%} & \textbf{6\%} & \textbf{3\%} & 6.63 & 0.15 & 0.97 \\
10 & \textbf{24\%} & \textbf{21\%} & \textbf{13\%} & \textbf{19\%} & -2\% & 6.07 & 0.16 & 0.96 \\
20 & 0\% & -4\% & \textbf{2\%} & \textbf{6\%} & \textbf{3\%} & 5.81 & 0.16 & 0.93 \\
30 & -1\% & -5\% & \textbf{7\%} & \textbf{17\%} & \textbf{4\%} & 6.13 & 0.16 & 0.95 \\
40 & \textbf{15\%} & \textbf{14\%} & \textbf{9\%} & \textbf{20\%} & 0\% & 7.47 & 0.13 & 0.95 \\
\midrule
\multicolumn{9}{l}{\textbf{Uniform}} \\
0 & \textbf{42\%} & \textbf{26\%} & \textbf{25\%} & \textbf{36\%} & \textbf{6\%} & 15.86 & 0.06 & 0.97 \\
10 & \textbf{27\%} & \textbf{20\%} & \textbf{14\%} & \textbf{30\%} & \textbf{2\%} & 9.42 & 0.11 & 1.00 \\
20 & \textbf{66\%} & \textbf{41\%} & \textbf{27\%} & \textbf{29\%} & -2\% & 10.02 & 0.10 & 0.95 \\
30 & \textbf{34\%} & \textbf{26\%} & \textbf{13\%} & \textbf{9\%} & 0\% & 12.50 & 0.08 & 0.96 \\
40 & \textbf{15\%} & \textbf{13\%} & \textbf{16\%} & \textbf{40\%} & -1\% & 13.66 & 0.07 & 0.97 \\
\midrule
\multicolumn{9}{l}{\textbf{Lognormal}} \\
0 & \textbf{1\%} & \textbf{2\%} & \textbf{4\%} & 0\% & -1\% & 5.41 & 0.18 & 0.97 \\
10 & \textbf{3\%} & \textbf{2\%} & \textbf{8\%} & 0\% & \textbf{1\%} & 5.23 & 0.18 & 0.96 \\
20 & \textbf{10\%} & \textbf{11\%} & \textbf{3\%} & -10\% & -2\% & 4.79 & 0.20 & 0.95 \\
30 & -12\% & -15\% & \textbf{4\%} & 0\% & \textbf{1\%} & 3.86 & 0.25 & 0.97 \\
40 & \textbf{4\%} & \textbf{3\%} & \textbf{2\%} & \textbf{4\%} & 0\% & 4.88 & 0.20 & 0.97 \\
\midrule
\multicolumn{9}{l}{\textbf{Gumbel}} \\
0 & \textbf{38\%} & \textbf{30\%} & \textbf{19\%} & \textbf{18\%} & -3\% & 6.90 & 0.14 & 0.96 \\
10 & \textbf{32\%} & \textbf{25\%} & \textbf{23\%} & \textbf{23\%} & -1\% & 4.56 & 0.21 & 0.96 \\
20 & -6\% & -8\% & -3\% & -5\% & \textbf{2\%} & 4.10 & 0.23 & 0.96 \\
30 & \textbf{13\%} & \textbf{12\%} & \textbf{3\%} & \textbf{8\%} & -1\% & 5.40 & 0.18 & 0.96 \\
40 & -3\% & -5\% & \textbf{10\%} & \textbf{10\%} & \textbf{1\%} & 5.14 & 0.19 & 0.96 \\
\bottomrule
\end{tabular}%
}

\end{small}
\caption{Test results for MultiMLPs, over 5 seeds, with quadratic $f(x)$ using 2 predictors, organized by distribution and seed. Relative metrics are in bold when positive.}
\label{tab:seeds-consolidated-multimlps-quadratic-x2}
\end{table}

\clearpage

\section{Full conformalized results on benchmark datasets}
\label{appendix:full-results}

\begin{table}[H]
\centering%
\begin{footnotesize}
\setlength{\tabcolsep}{4pt}
\resizebox{0.8\columnwidth}{!}{%
\begin{tabular}{lcccccccc}
\toprule

 & \multicolumn{5}{c}{Relative Metrics} & \multicolumn{3}{c}{Absolute Metrics} \\
\cmidrule(lr){2-6} \cmidrule(lr){7-9}
Dataset & $\Delta$CWR & $\Delta$PICP & $\Delta$NMPIW & $\Delta$CRPS & $\Delta$Pinball & CWR & NMPIW & PICP \\
\midrule

space\_ga & \textbf{9\%} & 0\% & \textbf{9\%} & \textbf{12\%} & \textbf{10\%} & 6.28 & 0.15 & 0.95 \\
space\_ga & \textbf{5\%} & \textbf{2\%} & \textbf{2\%} & \textbf{11\%} & \textbf{13\%} & 2.95 & 0.32 & 0.95 \\
space\_ga & \textbf{4\%} & \textbf{1\%} & \textbf{2\%} & \textbf{10\%} & \textbf{9\%} & 3.23 & 0.30 & 0.97 \\
space\_ga & \textbf{21\%} & -2\% & \textbf{19\%} & \textbf{11\%} & \textbf{9\%} & 3.37 & 0.28 & 0.96 \\
space\_ga & \textbf{16\%} & -2\% & \textbf{15\%} & \textbf{11\%} & \textbf{6\%} & 6.74 & 0.14 & 0.94 \\
cpu\_activity & \textbf{7\%} & 0\% & \textbf{7\%} & \textbf{17\%} & \textbf{10\%} & 9.84 & 0.10 & 0.96 \\
cpu\_activity & -3\% & \textbf{1\%} & -4\% & \textbf{10\%} & 0\% & 8.69 & 0.11 & 0.96 \\
cpu\_activity & \textbf{11\%} & -1\% & \textbf{11\%} & \textbf{14\%} & \textbf{10\%} & 10.20 & 0.09 & 0.94 \\
cpu\_activity & \textbf{11\%} & \textbf{1\%} & \textbf{10\%} & \textbf{26\%} & \textbf{36\%} & 10.14 & 0.09 & 0.95 \\
cpu\_activity & \textbf{1\%} & \textbf{1\%} & 0\% & \textbf{13\%} & \textbf{18\%} & 9.99 & 0.10 & 0.95 \\
naval\_propulsion\_plant & \textbf{200\%} & -3\% & \textbf{68\%} & \textbf{73\%} & \textbf{43\%} & 9.19 & 0.10 & 0.92 \\
naval\_propulsion\_plant & \textbf{114\%} & 0\% & \textbf{53\%} & \textbf{66\%} & \textbf{31\%} & 7.21 & 0.13 & 0.95 \\
naval\_propulsion\_plant & \textbf{159\%} & \textbf{1\%} & \textbf{61\%} & \textbf{74\%} & \textbf{50\%} & 7.97 & 0.12 & 0.96 \\
naval\_propulsion\_plant & \textbf{187\%} & 0\% & \textbf{65\%} & \textbf{74\%} & \textbf{50\%} & 8.73 & 0.11 & 0.95 \\
naval\_propulsion\_plant & \textbf{156\%} & \textbf{2\%} & \textbf{60\%} & \textbf{71\%} & \textbf{50\%} & 8.24 & 0.12 & 0.96 \\
miami\_housing & \textbf{9\%} & -1\% & \textbf{9\%} & \textbf{14\%} & -5\% & 9.08 & 0.10 & 0.95 \\
miami\_housing & -4\% & 0\% & -4\% & \textbf{8\%} & -12\% & 8.28 & 0.12 & 0.96 \\
miami\_housing & \textbf{2\%} & 0\% & \textbf{3\%} & \textbf{9\%} & -19\% & 8.49 & 0.11 & 0.94 \\
miami\_housing & \textbf{12\%} & 0\% & \textbf{10\%} & \textbf{17\%} & -5\% & 9.23 & 0.10 & 0.95 \\
miami\_housing & \textbf{8\%} & -1\% & \textbf{9\%} & \textbf{11\%} & -21\% & 8.70 & 0.11 & 0.95 \\
kin8nm & \textbf{16\%} & -1\% & \textbf{15\%} & \textbf{22\%} & \textbf{3\%} & 2.29 & 0.41 & 0.94 \\
kin8nm & \textbf{19\%} & -1\% & \textbf{16\%} & \textbf{25\%} & \textbf{11\%} & 2.31 & 0.41 & 0.94 \\
kin8nm & \textbf{16\%} & -1\% & \textbf{14\%} & \textbf{20\%} & \textbf{6\%} & 2.27 & 0.42 & 0.95 \\
kin8nm & \textbf{19\%} & -1\% & \textbf{17\%} & \textbf{22\%} & \textbf{8\%} & 2.29 & 0.41 & 0.95 \\
kin8nm & \textbf{19\%} & -1\% & \textbf{17\%} & \textbf{24\%} & \textbf{11\%} & 2.40 & 0.40 & 0.95 \\
concrete\_compressive\_strength & \textbf{31\%} & -1\% & \textbf{24\%} & \textbf{35\%} & \textbf{25\%} & 2.95 & 0.33 & 0.96 \\
concrete\_compressive\_strength & \textbf{73\%} & -5\% & \textbf{45\%} & \textbf{34\%} & \textbf{14\%} & 3.60 & 0.26 & 0.92 \\
concrete\_compressive\_strength & \textbf{27\%} & -3\% & \textbf{24\%} & \textbf{29\%} & \textbf{6\%} & 3.14 & 0.29 & 0.92 \\
concrete\_compressive\_strength & \textbf{47\%} & -2\% & \textbf{34\%} & \textbf{33\%} & \textbf{9\%} & 3.14 & 0.30 & 0.94 \\
concrete\_compressive\_strength & \textbf{24\%} & -1\% & \textbf{20\%} & \textbf{27\%} & \textbf{4\%} & 3.25 & 0.29 & 0.93 \\
cars & \textbf{32\%} & -2\% & \textbf{25\%} & \textbf{25\%} & \textbf{6\%} & 5.78 & 0.17 & 0.95 \\
cars & \textbf{10\%} & -1\% & \textbf{10\%} & \textbf{20\%} & \textbf{18\%} & 5.02 & 0.19 & 0.95 \\
cars & \textbf{27\%} & 0\% & \textbf{22\%} & \textbf{26\%} & \textbf{17\%} & 5.08 & 0.19 & 0.95 \\
cars & \textbf{67\%} & -3\% & \textbf{42\%} & \textbf{42\%} & \textbf{32\%} & 6.15 & 0.15 & 0.94 \\
cars & \textbf{22\%} & -2\% & \textbf{20\%} & \textbf{23\%} & \textbf{15\%} & 4.63 & 0.21 & 0.97 \\
energy\_efficiency & \textbf{114\%} & \textbf{1\%} & \textbf{52\%} & \textbf{66\%} & \textbf{43\%} & 15.29 & 0.06 & 0.96 \\
energy\_efficiency & \textbf{112\%} & 0\% & \textbf{53\%} & \textbf{67\%} & \textbf{50\%} & 14.18 & 0.07 & 0.95 \\
energy\_efficiency & \textbf{109\%} & -2\% & \textbf{53\%} & \textbf{64\%} & \textbf{44\%} & 10.78 & 0.09 & 0.96 \\
energy\_efficiency & \textbf{230\%} & \textbf{8\%} & \textbf{68\%} & \textbf{81\%} & \textbf{73\%} & 17.66 & 0.05 & 0.96 \\
energy\_efficiency & \textbf{72\%} & \textbf{5\%} & \textbf{39\%} & \textbf{59\%} & \textbf{58\%} & 11.84 & 0.08 & 0.95 \\
california\_housing & \textbf{3\%} & 0\% & \textbf{3\%} & \textbf{23\%} & -3\% & 2.09 & 0.46 & 0.96 \\
california\_housing & \textbf{4\%} & -1\% & \textbf{4\%} & \textbf{18\%} & -10\% & 2.27 & 0.42 & 0.94 \\
california\_housing & \textbf{2\%} & 0\% & \textbf{1\%} & \textbf{21\%} & -7\% & 2.29 & 0.41 & 0.95 \\
california\_housing & 0\% & 0\% & 0\% & \textbf{15\%} & -9\% & 2.27 & 0.41 & 0.94 \\
california\_housing & -1\% & 0\% & -1\% & \textbf{15\%} & -10\% & 2.28 & 0.42 & 0.95 \\
airfoil\_self\_noise & -5\% & 0\% & -5\% & \textbf{9\%} & -19\% & 2.60 & 0.37 & 0.96 \\
airfoil\_self\_noise & \textbf{6\%} & 0\% & \textbf{6\%} & \textbf{20\%} & \textbf{4\%} & 2.56 & 0.38 & 0.98 \\
airfoil\_self\_noise & -32\% & 0\% & -47\% & -36\% & -43\% & 1.57 & 0.63 & 0.99 \\
airfoil\_self\_noise & -4\% & 0\% & -5\% & \textbf{8\%} & -10\% & 2.26 & 0.43 & 0.97 \\
airfoil\_self\_noise & -8\% & -1\% & -7\% & \textbf{9\%} & -15\% & 2.43 & 0.39 & 0.95 \\
QSAR\_fish\_toxicity & \textbf{2\%} & 0\% & \textbf{1\%} & \textbf{8\%} & \textbf{10\%} & 1.80 & 0.54 & 0.97 \\
QSAR\_fish\_toxicity & \textbf{11\%} & -1\% & \textbf{10\%} & \textbf{8\%} & \textbf{12\%} & 1.63 & 0.59 & 0.96 \\
QSAR\_fish\_toxicity & \textbf{4\%} & \textbf{2\%} & \textbf{3\%} & \textbf{9\%} & \textbf{8\%} & 1.80 & 0.54 & 0.97 \\
QSAR\_fish\_toxicity & \textbf{26\%} & 0\% & \textbf{20\%} & \textbf{17\%} & \textbf{18\%} & 1.55 & 0.64 & 0.99 \\
QSAR\_fish\_toxicity & \textbf{53\%} & -1\% & \textbf{35\%} & \textbf{24\%} & \textbf{30\%} & 1.73 & 0.57 & 0.98 \\
\bottomrule
\end{tabular}%
}
\end{footnotesize}
\caption{Test results for MultiXGBs with complete data at 95\% quantiles for seeds 0, 10, 20, 30, 40, with rows ordered by seed (ascending). Performance over the baseline is highlighted in bold.}
\label{table-multixgbs-complete}
\end{table}

\begin{table}[H]
\centering%
\begin{footnotesize}
\setlength{\tabcolsep}{4pt}
\resizebox{0.8\columnwidth}{!}{%
\begin{tabular}{lcccccccc}
\toprule

 & \multicolumn{5}{c}{Relative Metrics} & \multicolumn{3}{c}{Absolute Metrics} \\
\cmidrule(lr){2-6} \cmidrule(lr){7-9}
Dataset & $\Delta$CWR & $\Delta$PICP & $\Delta$NMPIW & $\Delta$CRPS & $\Delta$Pinball & CWR & NMPIW & PICP \\
\midrule

space\_ga & \textbf{4\%} & -2\% & \textbf{6\%} & \textbf{6\%} & -9\% & 5.86 & 0.16 & 0.95 \\
space\_ga & -2\% & 0\% & -2\% & \textbf{7\%} & -3\% & 2.66 & 0.36 & 0.96 \\
space\_ga & \textbf{10\%} & -1\% & \textbf{10\%} & \textbf{7\%} & -5\% & 3.19 & 0.30 & 0.97 \\
space\_ga & -2\% & 0\% & -2\% & \textbf{7\%} & -5\% & 2.99 & 0.33 & 0.97 \\
space\_ga & \textbf{15\%} & -3\% & \textbf{16\%} & \textbf{11\%} & -9\% & 7.22 & 0.13 & 0.91 \\
cpu\_activity & \textbf{5\%} & -5\% & \textbf{10\%} & \textbf{2\%} & -22\% & 8.69 & 0.11 & 0.94 \\
cpu\_activity & \textbf{1\%} & -3\% & \textbf{4\%} & \textbf{7\%} & -11\% & 8.05 & 0.12 & 0.96 \\
cpu\_activity & \textbf{3\%} & -5\% & \textbf{8\%} & \textbf{5\%} & -11\% & 8.26 & 0.11 & 0.94 \\
cpu\_activity & -1\% & -4\% & \textbf{2\%} & -2\% & -22\% & 8.14 & 0.12 & 0.95 \\
cpu\_activity & \textbf{3\%} & -4\% & \textbf{7\%} & \textbf{2\%} & -22\% & 8.11 & 0.12 & 0.95 \\
naval\_propulsion\_plant & \textbf{154\%} & -6\% & \textbf{63\%} & \textbf{69\%} & \textbf{40\%} & 3.55 & 0.27 & 0.94 \\
naval\_propulsion\_plant & \textbf{153\%} & 0\% & \textbf{60\%} & \textbf{68\%} & \textbf{46\%} & 3.95 & 0.24 & 0.95 \\
naval\_propulsion\_plant & \textbf{160\%} & 0\% & \textbf{61\%} & \textbf{72\%} & \textbf{52\%} & 3.86 & 0.25 & 0.96 \\
naval\_propulsion\_plant & \textbf{136\%} & 0\% & \textbf{58\%} & \textbf{66\%} & \textbf{41\%} & 3.67 & 0.26 & 0.95 \\
naval\_propulsion\_plant & \textbf{118\%} & -4\% & \textbf{56\%} & \textbf{64\%} & \textbf{41\%} & 3.23 & 0.29 & 0.95 \\
miami\_housing & -13\% & \textbf{1\%} & -16\% & -4\% & -75\% & 6.25 & 0.15 & 0.96 \\
miami\_housing & -6\% & 0\% & -6\% & 0\% & -67\% & 6.71 & 0.14 & 0.95 \\
miami\_housing & -5\% & \textbf{1\%} & -6\% & 0\% & -67\% & 6.47 & 0.15 & 0.95 \\
miami\_housing & -5\% & 0\% & -4\% & -1\% & -75\% & 6.75 & 0.14 & 0.95 \\
miami\_housing & -15\% & \textbf{1\%} & -19\% & -7\% & -81\% & 6.03 & 0.16 & 0.95 \\
kin8nm & \textbf{8\%} & 0\% & \textbf{7\%} & \textbf{15\%} & 0\% & 2.14 & 0.44 & 0.95 \\
kin8nm & \textbf{7\%} & 0\% & \textbf{7\%} & \textbf{13\%} & -3\% & 2.11 & 0.45 & 0.95 \\
kin8nm & \textbf{7\%} & 0\% & \textbf{7\%} & \textbf{14\%} & 0\% & 2.10 & 0.46 & 0.96 \\
kin8nm & \textbf{5\%} & \textbf{1\%} & \textbf{4\%} & \textbf{12\%} & 0\% & 2.08 & 0.46 & 0.95 \\
kin8nm & \textbf{10\%} & \textbf{1\%} & \textbf{8\%} & \textbf{17\%} & 0\% & 2.22 & 0.43 & 0.95 \\
concrete\_compressive\_strength & \textbf{26\%} & -3\% & \textbf{23\%} & \textbf{12\%} & 0\% & 3.27 & 0.29 & 0.94 \\
concrete\_compressive\_strength & \textbf{5\%} & 0\% & \textbf{5\%} & \textbf{8\%} & 0\% & 2.97 & 0.33 & 0.97 \\
concrete\_compressive\_strength & \textbf{13\%} & \textbf{3\%} & \textbf{9\%} & \textbf{12\%} & \textbf{3\%} & 3.39 & 0.28 & 0.93 \\
concrete\_compressive\_strength & -14\% & \textbf{2\%} & -20\% & \textbf{4\%} & -12\% & 2.73 & 0.35 & 0.95 \\
concrete\_compressive\_strength & \textbf{3\%} & 0\% & \textbf{3\%} & \textbf{9\%} & -4\% & 2.87 & 0.34 & 0.97 \\
cars & \textbf{12\%} & -1\% & \textbf{12\%} & \textbf{3\%} & 0\% & 4.95 & 0.19 & 0.96 \\
cars & \textbf{2\%} & -2\% & \textbf{4\%} & -1\% & 0\% & 4.97 & 0.19 & 0.96 \\
cars & \textbf{29\%} & -3\% & \textbf{24\%} & \textbf{15\%} & \textbf{15\%} & 3.64 & 0.27 & 0.97 \\
cars & -25\% & \textbf{4\%} & -40\% & -8\% & -5\% & 6.19 & 0.15 & 0.92 \\
cars & \textbf{54\%} & -2\% & \textbf{36\%} & \textbf{20\%} & \textbf{28\%} & 4.70 & 0.21 & 0.97 \\
energy\_efficiency & \textbf{26\%} & -3\% & \textbf{23\%} & \textbf{12\%} & 0\% & 15.20 & 0.06 & 0.96 \\
energy\_efficiency & \textbf{16\%} & -3\% & \textbf{17\%} & \textbf{6\%} & 0\% & 16.14 & 0.06 & 0.94 \\
energy\_efficiency & \textbf{24\%} & 0\% & \textbf{20\%} & \textbf{9\%} & \textbf{25\%} & 14.43 & 0.07 & 0.99 \\
energy\_efficiency & \textbf{8\%} & 0\% & \textbf{8\%} & \textbf{4\%} & 0\% & 20.70 & 0.04 & 0.91 \\
energy\_efficiency & \textbf{1\%} & -1\% & \textbf{1\%} & -15\% & -80\% & 13.19 & 0.07 & 0.96 \\
california\_housing & -11\% & -1\% & -11\% & \textbf{16\%} & -26\% & 1.55 & 0.62 & 0.96 \\
california\_housing & -1\% & -2\% & \textbf{1\%} & \textbf{20\%} & -24\% & 1.68 & 0.57 & 0.95 \\
california\_housing & \textbf{2\%} & -2\% & \textbf{4\%} & \textbf{24\%} & -17\% & 1.70 & 0.56 & 0.95 \\
california\_housing & \textbf{4\%} & -3\% & \textbf{7\%} & \textbf{20\%} & -25\% & 1.76 & 0.53 & 0.93 \\
california\_housing & \textbf{9\%} & -3\% & \textbf{11\%} & \textbf{25\%} & -17\% & 1.83 & 0.52 & 0.94 \\
airfoil\_self\_noise & -21\% & -1\% & -24\% & -26\% & -40\% & 2.15 & 0.45 & 0.97 \\
airfoil\_self\_noise & \textbf{22\%} & 0\% & \textbf{18\%} & \textbf{19\%} & \textbf{18\%} & 2.91 & 0.34 & 0.98 \\
airfoil\_self\_noise & -20\% & -2\% & -22\% & -21\% & -36\% & 2.17 & 0.44 & 0.96 \\
airfoil\_self\_noise & -39\% & \textbf{1\%} & -66\% & -52\% & -61\% & 1.52 & 0.64 & 0.98 \\
airfoil\_self\_noise & -6\% & -1\% & -5\% & -3\% & -12\% & 2.49 & 0.39 & 0.97 \\
QSAR\_fish\_toxicity & -13\% & 0\% & -14\% & -4\% & -10\% & 1.69 & 0.57 & 0.97 \\
QSAR\_fish\_toxicity & -20\% & 0\% & -26\% & -16\% & -19\% & 1.36 & 0.72 & 0.98 \\
QSAR\_fish\_toxicity & \textbf{17\%} & -4\% & \textbf{18\%} & \textbf{6\%} & -2\% & 2.00 & 0.47 & 0.94 \\
QSAR\_fish\_toxicity & \textbf{17\%} & -2\% & \textbf{16\%} & \textbf{4\%} & \textbf{2\%} & 1.87 & 0.52 & 0.97 \\
QSAR\_fish\_toxicity & -18\% & 0\% & -22\% & -7\% & -14\% & 1.60 & 0.61 & 0.97 \\
\bottomrule
\end{tabular}%
}
\end{footnotesize}
\caption{Test results for MultiETs with complete data at 95\% quantiles for seeds 0, 10, 20, 30, 40, with rows ordered by seed (ascending). Performance over the baseline is highlighted in bold.}
\label{table-multiets-complete}
\end{table}

\begin{table}[H]
\centering%
\begin{footnotesize}
\setlength{\tabcolsep}{4pt}
\resizebox{0.8\columnwidth}{!}{%
\begin{tabular}{lcccccccc}
\toprule

 & \multicolumn{5}{c}{Relative Metrics} & \multicolumn{3}{c}{Absolute Metrics} \\
\cmidrule(lr){2-6} \cmidrule(lr){7-9}
Dataset & $\Delta$CWR & $\Delta$PICP & $\Delta$NMPIW & $\Delta$CRPS & $\Delta$Pinball & CWR & NMPIW & PICP \\
\midrule

space\_ga & \textbf{5\%} & -3\% & \textbf{8\%} & 0\% & -8\% & 7.12 & 0.13 & 0.92 \\
space\_ga & -6\% & -2\% & -4\% & -9\% & -13\% & 3.01 & 0.32 & 0.95 \\
space\_ga & \textbf{3\%} & 0\% & \textbf{3\%} & \textbf{2\%} & \textbf{3\%} & 3.60 & 0.27 & 0.97 \\
space\_ga & \textbf{25\%} & -2\% & \textbf{21\%} & \textbf{13\%} & \textbf{8\%} & 4.46 & 0.21 & 0.93 \\
space\_ga & 0\% & 0\% & -1\% & \textbf{3\%} & \textbf{8\%} & 7.29 & 0.13 & 0.94 \\
cpu\_activity & -16\% & -1\% & -18\% & -14\% & -10\% & 8.51 & 0.11 & 0.95 \\
cpu\_activity & -9\% & 0\% & -11\% & -9\% & -11\% & 8.44 & 0.11 & 0.96 \\
cpu\_activity & -14\% & 0\% & -16\% & -11\% & -11\% & 8.43 & 0.11 & 0.96 \\
cpu\_activity & -16\% & \textbf{1\%} & -20\% & -1\% & \textbf{17\%} & 8.73 & 0.11 & 0.95 \\
cpu\_activity & -14\% & 0\% & -16\% & -8\% & 0\% & 8.75 & 0.11 & 0.95 \\
naval\_propulsion\_plant & \textbf{84\%} & -1\% & \textbf{46\%} & \textbf{46\%} & \textbf{33\%} & 15.95 & 0.06 & 0.94 \\
naval\_propulsion\_plant & \textbf{26\%} & 0\% & \textbf{20\%} & \textbf{27\%} & 0\% & 12.02 & 0.08 & 0.95 \\
naval\_propulsion\_plant & -2\% & -1\% & -1\% & -3\% & -25\% & 13.52 & 0.07 & 0.94 \\
naval\_propulsion\_plant & \textbf{8\%} & \textbf{1\%} & \textbf{7\%} & \textbf{14\%} & -25\% & 11.94 & 0.08 & 0.96 \\
naval\_propulsion\_plant & \textbf{1\%} & 0\% & \textbf{1\%} & \textbf{2\%} & -20\% & 10.46 & 0.09 & 0.95 \\
miami\_housing & -19\% & -1\% & -23\% & -13\% & -35\% & 9.00 & 0.11 & 0.95 \\
miami\_housing & -35\% & 0\% & -56\% & -23\% & -38\% & 7.03 & 0.14 & 0.96 \\
miami\_housing & -13\% & 0\% & -15\% & -9\% & -33\% & 9.98 & 0.09 & 0.94 \\
miami\_housing & -20\% & 0\% & -25\% & -9\% & -11\% & 9.56 & 0.10 & 0.94 \\
miami\_housing & -17\% & 0\% & -21\% & -9\% & -12\% & 9.71 & 0.10 & 0.95 \\
kin8nm & \textbf{17\%} & 0\% & \textbf{14\%} & \textbf{14\%} & \textbf{15\%} & 4.50 & 0.21 & 0.96 \\
kin8nm & \textbf{4\%} & \textbf{1\%} & \textbf{3\%} & \textbf{5\%} & \textbf{6\%} & 4.86 & 0.20 & 0.95 \\
kin8nm & \textbf{18\%} & -2\% & \textbf{16\%} & \textbf{11\%} & \textbf{11\%} & 4.97 & 0.19 & 0.93 \\
kin8nm & \textbf{2\%} & 0\% & \textbf{2\%} & \textbf{5\%} & \textbf{6\%} & 4.59 & 0.21 & 0.95 \\
kin8nm & \textbf{3\%} & 0\% & \textbf{3\%} & \textbf{5\%} & 0\% & 4.80 & 0.20 & 0.93 \\
concrete\_compressive\_strength & \textbf{54\%} & -4\% & \textbf{37\%} & \textbf{26\%} & \textbf{21\%} & 3.33 & 0.29 & 0.96 \\
concrete\_compressive\_strength & -5\% & -2\% & -2\% & -2\% & -19\% & 3.15 & 0.30 & 0.94 \\
concrete\_compressive\_strength & \textbf{13\%} & -3\% & \textbf{14\%} & \textbf{8\%} & -12\% & 3.51 & 0.26 & 0.92 \\
concrete\_compressive\_strength & \textbf{73\%} & -2\% & \textbf{43\%} & \textbf{35\%} & \textbf{31\%} & 3.47 & 0.28 & 0.97 \\
concrete\_compressive\_strength & \textbf{19\%} & \textbf{2\%} & \textbf{14\%} & \textbf{17\%} & \textbf{15\%} & 3.67 & 0.26 & 0.95 \\
cars & -11\% & -2\% & -11\% & -13\% & -29\% & 5.13 & 0.18 & 0.94 \\
cars & -7\% & -1\% & -7\% & -9\% & -13\% & 5.21 & 0.19 & 0.97 \\
cars & -18\% & 0\% & -22\% & -17\% & -19\% & 4.00 & 0.24 & 0.98 \\
cars & \textbf{21\%} & -5\% & \textbf{22\%} & \textbf{5\%} & -21\% & 7.75 & 0.12 & 0.92 \\
cars & -44\% & \textbf{7\%} & -89\% & -23\% & -14\% & 4.89 & 0.20 & 0.99 \\
energy\_efficiency & \textbf{102\%} & \textbf{5\%} & \textbf{48\%} & \textbf{56\%} & \textbf{50\%} & 19.46 & 0.05 & 0.96 \\
energy\_efficiency & \textbf{74\%} & -1\% & \textbf{42\%} & \textbf{40\%} & \textbf{40\%} & 19.62 & 0.05 & 0.95 \\
energy\_efficiency & \textbf{65\%} & -3\% & \textbf{41\%} & \textbf{36\%} & \textbf{25\%} & 18.91 & 0.05 & 0.96 \\
energy\_efficiency & \textbf{46\%} & \textbf{3\%} & \textbf{30\%} & \textbf{33\%} & 0\% & 26.91 & 0.03 & 0.90 \\
energy\_efficiency & \textbf{149\%} & -6\% & \textbf{63\%} & \textbf{42\%} & 0\% & 24.65 & 0.04 & 0.92 \\
california\_housing & -10\% & 0\% & -11\% & 0\% & -3\% & 1.99 & 0.48 & 0.96 \\
california\_housing & -14\% & 0\% & -16\% & -5\% & -13\% & 2.12 & 0.45 & 0.94 \\
california\_housing & -11\% & 0\% & -12\% & -1\% & -9\% & 2.12 & 0.45 & 0.95 \\
california\_housing & -7\% & -1\% & -6\% & 0\% & -9\% & 2.22 & 0.42 & 0.94 \\
california\_housing & -12\% & 0\% & -14\% & -2\% & -9\% & 2.20 & 0.43 & 0.95 \\
airfoil\_self\_noise & \textbf{60\%} & -1\% & \textbf{38\%} & \textbf{33\%} & \textbf{30\%} & 5.45 & 0.18 & 0.97 \\
airfoil\_self\_noise & \textbf{55\%} & -1\% & \textbf{36\%} & \textbf{37\%} & \textbf{27\%} & 4.83 & 0.20 & 0.97 \\
airfoil\_self\_noise & \textbf{63\%} & 0\% & \textbf{39\%} & \textbf{43\%} & \textbf{32\%} & 4.98 & 0.20 & 0.98 \\
airfoil\_self\_noise & \textbf{17\%} & \textbf{1\%} & \textbf{14\%} & \textbf{20\%} & \textbf{23\%} & 4.84 & 0.20 & 0.95 \\
airfoil\_self\_noise & \textbf{16\%} & -2\% & \textbf{15\%} & \textbf{17\%} & \textbf{5\%} & 4.48 & 0.21 & 0.95 \\
QSAR\_fish\_toxicity & \textbf{21\%} & -3\% & \textbf{19\%} & \textbf{4\%} & \textbf{6\%} & 1.85 & 0.52 & 0.96 \\
QSAR\_fish\_toxicity & \textbf{2\%} & \textbf{1\%} & \textbf{1\%} & \textbf{1\%} & \textbf{2\%} & 1.57 & 0.62 & 0.98 \\
QSAR\_fish\_toxicity & -2\% & 0\% & -2\% & -3\% & \textbf{2\%} & 1.67 & 0.58 & 0.97 \\
QSAR\_fish\_toxicity & \textbf{36\%} & -3\% & \textbf{28\%} & \textbf{13\%} & \textbf{15\%} & 2.04 & 0.47 & 0.96 \\
QSAR\_fish\_toxicity & -15\% & \textbf{1\%} & -19\% & -9\% & -11\% & 1.43 & 0.69 & 0.98 \\
\bottomrule
\end{tabular}%
}
\end{footnotesize}
\caption{Test results for MultiMLPs with complete data at 95\% quantiles for seeds 0, 10, 20, 30, 40, with rows ordered by seed (ascending). Performance over the baseline is highlighted in bold.}
\label{table-multimlps-complete}
\end{table}

\end{document}